\documentclass{article}
\usepackage{style}
\usepackage{times}
\usepackage{soul}
\usepackage{url}
\usepackage[colorlinks=true, urlcolor=blue, linkcolor=blue, citecolor=black]{hyperref}
\usepackage[utf8]{inputenc}
\usepackage[small]{caption}
\usepackage{graphicx}
\usepackage{amsmath}
\usepackage{amsthm}
\usepackage{booktabs}
\usepackage{algorithm}
\usepackage{algorithmic}
\usepackage{tikz}
\usetikzlibrary{shapes.geometric, arrows.meta, positioning}

\usepackage{natbib}
\setcitestyle{round}

\title{Can Small Language Models Be Fine-Tuned to Reliably Reason about Cardinal Directions?}

\author{
    Dimitrios Andreakis${^1}$\and
    Robert E. Blackwell${^{2}}$\and
    Anthony G. Cohn${^{2}}$\And
    Manolis Koubarakis${^1}$\\
    \affiliations
    $^1$Department of Informatics and Telecommunications, National and Kapodistrian University of Athens, Greece\\
    $^2$School of Computer Science, University of Leeds, UK\\
    \emails
    \href{mailto:sdi2300008@di.uoa.gr}{sdi2300008@di.uoa.gr},
    \href{mailto:r.e.blackwell@leeds.ac.uk}{r.e.blackwell@leeds.ac.uk},
    \href{mailto:A.G.Cohn@leeds.ac.uk}{A.G.Cohn@leeds.ac.uk},
    \href{mailto:koubarak@di.uoa.gr}{koubarak@di.uoa.gr}
}

\newcommand{\bench}{COSIT-CD}

\begin{document}

\maketitle

\begin{abstract}
\citet{cohn2024evaluating,cohn2025evaluating} found that small language models achieve substantially lower accuracy than large language models on cardinal direction reasoning tasks. This capability gap poses a substantial barrier to deploying spatially-aware autonomous agents on resource-constrained edge devices (e.g., real-time embodied robotics and local navigation agents). In this paper, we investigate whether small language models can be fine-tuned to reliably perform cardinal direction reasoning. Through a controlled, 11-part diagnostic ablation study on the benchmark of \citet{cohn2024evaluating} (hereafter referred to as the \bench{} benchmark), we evaluate a 1-billion-parameter model (\texttt{gemma-3-1b-it}) to reverse-engineer its spatial learning mechanics. We discover that text-based small language models do not possess an inherent capacity for cardinal direction reasoning; rather, their spatial deduction relies heavily on discrete syntactic pattern matching. We demonstrate that while In-Context Learning and generic spatial datasets (StepGame) fail to bridge this reasoning gap, targeted weight updates can teach the model to correctly apply conflicting spatial rules and ignore irrelevant vocabulary. Ultimately, we define the ``Minimum Viable Dataset'', a rigorously curated framework designed to isolate the exact syntactic thresholds required for cardinal direction reasoning. By explicitly mapping these syntactic structures, our model trained on this dataset achieves 100.00\% accuracy on the \bench{} benchmark, demonstrating that reliable, domain-specific spatial logic can be embedded directly into highly constrained architectures. All scripts, datasets, and experimental results are publicly available at \href{https://github.com/EncodedMind/Can-Small-Language-Models-reason-about-cardinal-directions}{our GitHub repository}.
\end{abstract}

\section{Introduction}

The deployment of autonomous systems, such as real-time embodied robotics and local navigation agents, is fundamentally bottlenecked by hardware constraints \citep{firoozi2023foundation, zhou2019edge}. While State-of-the-Art (SOTA) Large Language Models (LLMs) and Large Reasoning Models (LRMs) possess immense generalized knowledge, running these massive architectures at production scale entails substantial infrastructure investments, high latencies, and heavy computational requirements. This operational reality motivates an essential shift toward Small Language Models (SLMs) that can perform specialized reasoning directly on devices with limited computational capabilities, ranging from mobile phones and laptops to embodied edge-AI systems like autonomous vehicles and robotics.

However, a core requirement for these embodied systems is the ability to reliably navigate and reason about physical space. Recent evaluations by \citet{cohn2024evaluating,cohn2025evaluating} assessed whether LLMs can accurately reason about cardinal (e.g., North, East) and intercardinal (e.g., North-East, South-West) directions from text-based spatial scenarios. Their findings highlighted a key limitation: while high-parameter LRMs still struggle to achieve perfect reliability on complex spatial reasoning benchmarks, smaller models achieve substantially lower accuracy in these evaluations. This performance gap raises a central research question: \textit{Can Small Language Models be fine-tuned to achieve SOTA performance in specialized domains?}

Training an SLM's reasoning capabilities introduces the severe risk of shortcut learning. If a model is merely trained on spatial data without strict syntactic structure, it will memorize localized vocabulary rather than learning the underlying spatial logic. Therefore, in this work, we do not merely optimize an SLM for a high benchmark score; we systematically reverse-engineer \textit{how} text-based models process text describing spatial scenarios. 

Using the \texttt{gemma-3-1b-it} model \citep{gemma2025gemma3}, we establish a zero-shot baseline of 4.38\% on the ``Large'' benchmark of \citet{cohn2024evaluating} (hereafter referred to as the \bench{} benchmark). This benchmark evaluates six distinct templates that require models to infer implicit spatial relationships from text (e.g., ``I am running east along the north shore of the lake. In which direction is the lake?'' $\rightarrow$ South). We demonstrate that exposing the model to generic multi-hop spatial graphs (e.g., StepGame) or prompt engineering (In-Context Learning) yields only marginal improvements, failing to resolve the performance gap with frontier LLMs. Through a rigorous diagnostic ablation pipeline, we provide strong empirical evidence that the SLM acts strictly as a linguistic parser. It does not appear to construct true spatial representations for cardinal direction reasoning; rather, its spatial reasoning relies on matching syntactic patterns, a heuristic that breaks down when the model encounters unfamiliar vocabulary.

By identifying the precise limitations of the model's pattern-matching behavior, we propose the ``Minimum Viable Dataset'' (MVD) framework. We demonstrate that by fine-tuning the model on a curated dataset designed to resolve conflicting spatial rules (such as distinguishing between inner and outer objects) while minimizing irrelevant vocabulary, we can reliably embed task-specific cardinal direction reasoning capabilities into the model. Ultimately, our fine-tuned 1B model achieves 100.00\% accuracy across all benchmark templates, entirely outperforming generalist models hundreds of times its size on this specific task, while retaining an acceptable level of general language capabilities for edge deployment.

The remainder of this paper is structured as follows: Section 2 reviews related work in spatial reasoning and shortcut learning. Section 3 details our diagnostic methodology and the \bench{} benchmark setup. Section 4 presents the initial diagnostic ablation experiments mapping the SLM's failure modes. Sections 5 and 6 analyze the failures of alternative approaches, specifically In-Context Learning and the limitations of generic spatial datasets. Sections 7 through 9 define the structural rules of the MVD framework and detail its iterative syntactic engineering and final evaluation. Section 10 demonstrates the framework's architecture agnosticism across additional model families. Finally, Sections 11 through 13 contextualize our findings, comparing the SLM's pattern matching against true reasoning, contrasting our 100\% accuracy against massive LRMs, and outlining future work. General intelligence retention and example MVD prompts are provided in Appendices A and B, respectively.

\section{Background and Related Work}

\subsection{Evaluating Spatial Reasoning in LLMs}
Qualitative Spatial Reasoning is a vital capability for autonomous navigation and embodied artificial intelligence, yet language models consistently struggle in this domain. \cite{yamada2024evaluating} and \cite{wularge} established that foundational LLMs, such as GPT-4 \citep{openai2023gpt4}, LLaMA 2 \citep{touvron2023llama2}, and Gemini \citep{geminiteam2023gemini}, struggle to construct accurate representations of spatial structures, often scoring near random chance on directional tasks. \cite{cohn2024evaluating} expanded this by introducing a cardinal direction reasoning benchmark (the \bench{} benchmark) to evaluate models on their ability to deduce cardinal directions from textual scenarios based on distinct topological templates. In a subsequent evaluation, they demonstrated that even multi-billion-parameter LRMs struggle to generalize across linguistic variations \citep{cohn2025evaluating}. 

\cite{gardelakos2025can} recently corroborated this, showing that despite immense inference-time compute, LRMs still fail to perfectly compute the composition of the cardinal direction calculus CDC \citep{goyal1997, goyal2001, skiadopoulos2004}. Furthermore, \cite{xie2025evaluating} revealed that topological entanglement (e.g., distinguishing inner versus outer objects) remains a fundamental weakness in current NLP architectures. \cite{yin2025multimodal} demonstrated that even Multimodal Large Language Models (MLLMs) fail to reliably understand compass directions, emphasizing the depth of this cognitive gap across the studied modalities.

\subsection{The Efficacy of Small Language Models}
The high operational costs of LLMs have motivated research into SLMs, especially for specialized edge-computing workflows. \cite{belcak2025small} argue that SLMs are more economical than LLMs while still being suitable for agentic systems, where tasks are typically narrow and repetitive. This was empirically validated by \cite{jhandi2025small}, who demonstrated that task-specific structural optimization fundamentally outperforms scale-based generalization, allowing models with only millions of parameters to outperform much larger models on structured tasks. These findings suggest that SLMs can avoid spreading their limited capacity across broad, generalized tasks by focusing entirely on structured patterns.

\subsection{Shortcut Learning}
Despite the promise of SLMs, evaluating their true capabilities is complicated by the phenomenon of \textit{shortcut learning}. As defined by \cite{du2022shortcut}, language models frequently exploit dataset biases and superficial syntactic artifacts to solve natural language understanding (NLU) tasks, rather than acquiring robust, generalizable logic. To overcome these syntactic dependencies, standard Supervised Fine-Tuning (SFT) is often insufficient. Recent literature emphasizes the necessity of guided data generation and reinforcement learning techniques to build capable domain-specific SLMs. For example, \cite{kumar2025building} demonstrated that integrating Direct Preference Optimization (DPO) significantly enhances an SLM's domain generalization and robustness. 

\subsection{The Gap: SLMs in Spatial Reasoning}
Consequently, it remains unknown whether targeted fine-tuning can solve the spatial reasoning deficit without falling prey to shortcut learning. \cite{burnell2023rethink} argues that the AI community must move beyond reporting aggregate benchmark scores, which often obscure underlying biases and shortcut heuristics. Aligning with this directive, our primary objective is not merely to report high accuracy on the \bench{} dataset. Instead, we systematically ablate the SLM's training data to gain deep insight into its susceptibility to shortcut learning, identifying the exact threshold where shortcut learning ends and robust syntactic mapping begins.

\section{Diagnostic Probes and Experimental Setup}

To determine whether SLMs can develop genuine spatial cognition rather than relying on shallow linguistic heuristics, we require a highly controlled testing environment. Rather than training the model on a generalized spatial dataset immediately, we utilize targeted subsets of the \bench{} benchmark. This approach is deliberately designed as a diagnostic probe. By isolating specific syntactic structures and exposing the model to intentional data imbalances, we purposefully trigger and observe failure modes such as shortcut learning and lexical entanglement. These diagnostics are essential for mapping the model's limitations and establishing the strict structural requirements for the Minimum Viable Dataset (MVD) introduced later in this work.

\subsection{The \bench{} Diagnostic Framework}

To evaluate spatial reasoning systematically, this study utilizes the ``Large'' dataset of \citet{cohn2024evaluating}. The dataset comprises 5,760 questions algorithmically generated to test a model's ability to deduce cardinal and intercardinal directions within realistic locomotion scenarios. The complete dataset of questions and ground-truth answers is publicly available via the original authors' \href{https://github.com/alan-turing-institute/cosit-2024-evaluating-the-ability-of-llms-to-reason-about-cardinal-directions/tree/main/data/cosit24/large}{GitHub repository}.

The dataset is constructed from six distinct syntactic templates (T1 through T6). Each template embeds a specific spatial rule governing the relationship between the actor's spatial vector and a target geographical object. The templates, with placeholder variables bracketed, are defined as follows:

\begin{itemize}
    \item \textbf{Template 1 (T1) (Inner Object):} You are walking [south] along the [east] shore of a lake. In which direction is the lake?
    \item \textbf{Template 2 (T2) (Inner Object + Vector Reversal):} You are walking [south] along the [east] shore of a lake and then turn around to head back in the direction you came from, in which direction is the lake?
    \item \textbf{Template 3 (T3) (Inner Object, Alternative Noun):} You are walking [south] along the middle of the [east] side of a park; in which direction is the bandstand located in the center of the park?
    \item \textbf{Template 4 (T4) (Linear Object):} You are walking [east] along the [south] side of a road which runs [east to west]. In which direction is the road?
    \item \textbf{Template 5 (T5) (Outer Object):} You are walking [south] along the [east] shore of the island. In which direction is the sea?
    \item \textbf{Template 6 (T6) (Outer Object + Vector Reversal):} You are walking [south] along the [east] shore of an island and then turn around to head back in the direction you came from, in which direction is the sea?
\end{itemize}

\begin{figure*}[ht]
    \centering
    \begin{tikzpicture}[
        >=stealth,
        font=\sffamily,
        water/.style={fill=cyan!15, draw=cyan!60, thick, rounded corners=10pt},
        land/.style={fill=brown!15, draw=brown!60, thick, rounded corners=10pt},
        person/.style={circle, fill=red!80, inner sep=2.5pt},
        walkarrow/.style={->, >={Stealth[scale=1.2]}, red!80, ultra thick},
        lookarrow/.style={->, >={Stealth[scale=1.2]}, blue!70, dashed, ultra thick}
    ]

    % ==========================================
    % LEFT PANEL: TEMPLATE 1 (LAKE)
    % ==========================================
    \begin{scope}[shift={(-5.5,0)}]
        % Background (Land)
        \draw[land, rounded corners=0pt] (-3,-3) rectangle (3,3);
        % Foreground (Lake)
        \draw[water] (-2,-2) rectangle (2,2);
        
        \node[font=\Large\bfseries, text=cyan!80!black] at (0, 0.6) {LAKE};
        \node[font=\bfseries] at (0, 3.5) {Template 1: Enclosed Water};
        
        % The Person on the East Shore
        \node[person, label={[font=\bfseries]right:Person}] (p1) at (2, 1) {};
        
        % Walking Vector
        \draw[walkarrow] (p1) -- (2, -1.5) node[midway, right, align=left] {Walking\\South};
        
        % Target Vector
        \draw[lookarrow] (1.6, -0.8) -- (0.0, -0.8) node[midway, above, text=blue!70!black] {Water is West};
        
        % Lexical annotation
        \node[fill=white, draw=black, rounded corners=2pt, inner sep=4pt, text width=4.5cm, align=center] at (0, -4.2) {
            \textbf{Lexical Cue:} ``East shore''\\
            \textbf{Directional Target:} West
        };
    \end{scope}

    % ==========================================
    % RIGHT PANEL: TEMPLATE 5 (ISLAND)
    % ==========================================
    \begin{scope}[shift={(5.5,0)}]
        % Background (Sea)
        \draw[water, rounded corners=0pt] (-3,-3) rectangle (3,3);
        % Foreground (Island)
        \draw[land] (-2,-2) rectangle (2,2);
        
        \node[font=\Large\bfseries, text=brown!80!black] at (0, 0.6) {ISLAND};
        \node[font=\bfseries] at (0, 3.5) {Template 5: Enclosed Land};
        
        % The Person on the East Shore
        \node[person, label={[font=\bfseries]left:Person}] (p2) at (2, 1) {};
        
        % Walking Vector
        \draw[walkarrow] (p2) -- (2, -1.5) node[midway, left, align=right] {Walking\\South};
        
        % Target Vector
        \draw[lookarrow] (2.3, -0.8) -- (3.9, -0.8) node[midway, above, text=blue!70!black] {Water is East};
        
        % Lexical annotation
        \node[fill=white, draw=black, rounded corners=2pt, inner sep=4pt, text width=4.5cm, align=center] at (0, -4.2) {
            \textbf{Lexical Cue:} ``East shore''\\
            \textbf{Directional Target:} East
        };
    \end{scope}

    % ==========================================
    % COMPASS ROSE (Center)
    % ==========================================
    \begin{scope}[shift={(0,0)}]
        \draw[thick, <->] (-1.2,0) node[left, font=\bfseries] {W} -- (1.2,0) node[right, font=\bfseries] {E};
        \draw[thick, <->] (0,-1.2) node[below, font=\bfseries] {S} -- (0,1.2) node[above, font=\bfseries] {N};
    \end{scope}

    \end{tikzpicture}
    \caption{Visualizing Lexical Ambiguity. Despite identical lexical cues, the inverted structural boundaries of a lake versus an island require mathematically opposite spatial vectors: walking South on the East shore of a lake means the water is West, whereas walking South on the East shore of an island means the water is East. This inherent conflict demonstrates why models must be trained to compute the underlying spatial relationship rather than defaulting to shallow lexical pattern matching.}
    \label{fig:structural_conflict}
\end{figure*}
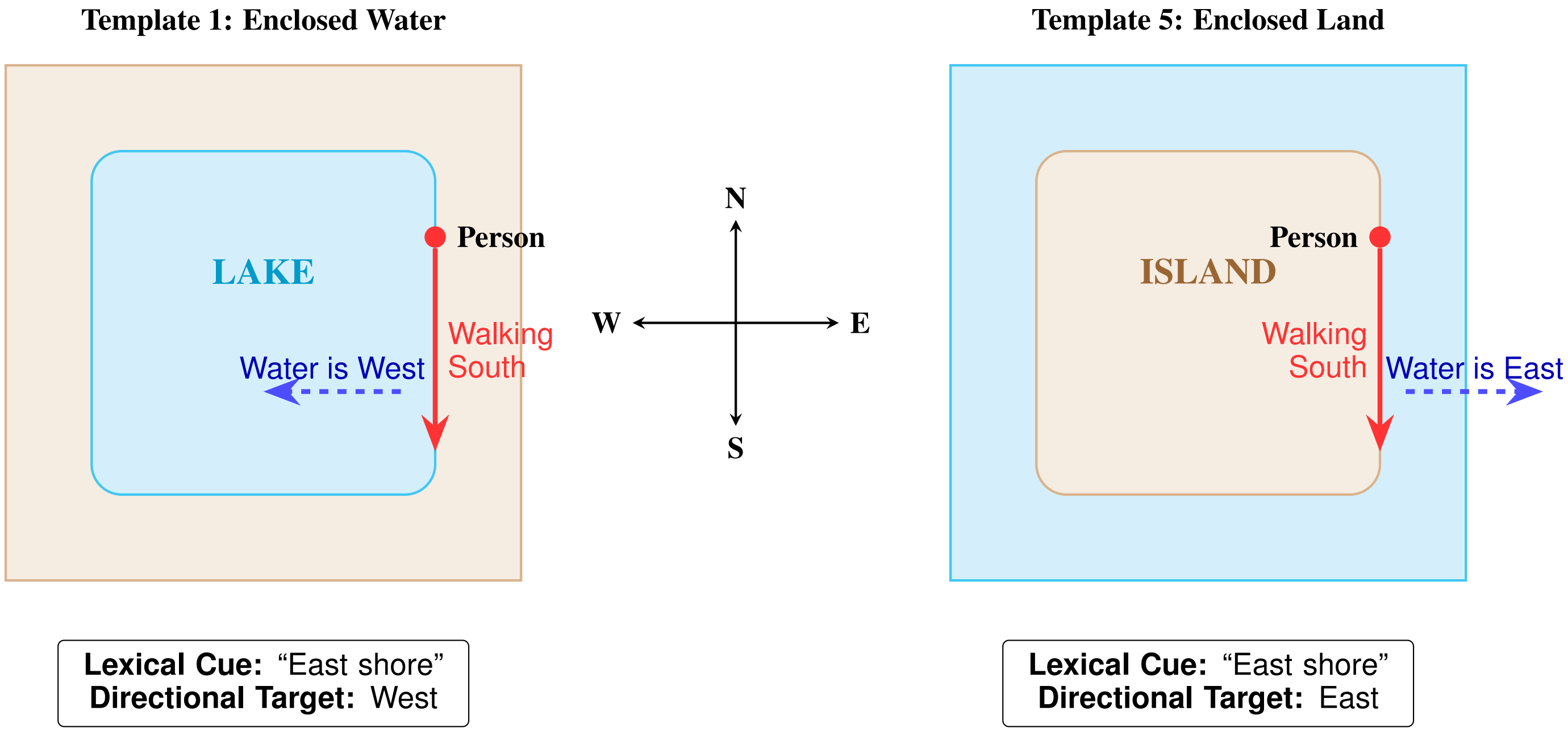

These templates allow for extensive variation. The bracketed variables iterate through all eight cardinal and intercardinal directions. Note that once the first direction vector is established in a template, there are strictly two valid spatial possibilities for the second direction. Furthermore, to prevent reliance on specific text patterns, the dataset applies superficial linguistic variations to each template. Specifically, it cycles through ten distinct forms of locomotion (\textit{cycling, driving, hiking, jogging, perambulating, racing, riding, running, unicycling, walking}) and six grammatical person forms (\textit{``I am,'' ``We are,'' ``You are,'' ``He is,'' ``She is,'' ``They are''}). This exhaustive combination of variables (6 templates $\times$ 10 locomotion forms $\times$ 6 person forms $\times$ 8 initial directions $\times$ 2 secondary directions) yields the complete dataset of 5,760 unique questions.

Templates T1/T2 and T5/T6 introduce intentional structural conflict. As illustrated in Figure \ref{fig:structural_conflict}, while they share the ambiguous keyword ``shore,'' they require opposite spatial calculations (the lake is internal to the shore, while the sea is external to the island).

During evaluation, zero-shot prompting was utilized. To constrain the model's output and prevent verbose explanations, we adopted the same system prompt from \citet{cohn2024evaluating} across all experiments:

\begin{quote}
    \textit{``You are a helpful assistant. I will give you a question about directions. The answer is either north, south, east, west, north-east, north-west, south-east or south-west. Please only reply with the answer. No yapping.''}
\end{quote}

\subsection{Model and Training Configuration}

\paragraph{Hardware and Model Selection.} 
A primary motivation of this research is the deployment of spatial reasoning capabilities on resource-constrained edge-AI systems. All fine-tuning and evaluation experiments were conducted on a consumer-grade workstation equipped with an Intel Core i5-11400F CPU and a single NVIDIA GeForce RTX 2060 GPU (12GB VRAM). To satisfy the strict memory constraints of edge deployment while ensuring robust instruction-following capabilities and reasonable inference speeds,
the 1-billion-parameter Gemma 3 Instruct model (\texttt{gemma-3-1b-it}) was selected as a representative architecture (alongside comparable alternatives such as Llama~3.2 and Qwen~3.5, evaluated in Section~\ref{sec:agnosticism}).

\paragraph{Hyperparameter Optimization.} 
We performed full-weight fine-tuning (updating all model parameters rather than using parameter-efficient methods such as LoRA \citep{hu2022lora}). Prior to the dataset ablation study, a grid search was conducted to identify the optimal training hyperparameters for the spatial reasoning task. The optimization space included variations in learning rate ($1\times10^{-5}$, $3\times10^{-5}$, $5\times10^{-5}$), batch size to gradient accumulation ratios (1:8, 2:4, and 4:2), and total training epochs (3, 5, and 10). The baseline configuration that successfully achieved 100\% accuracy utilized a learning rate of $1\times10^{-5}$, a per-device batch size of 4 with 2 gradient accumulation steps (yielding an effective batch size of 8), trained over 5 epochs. To ensure a controlled ablation study where performance variations reflect only the withheld data variables, this baseline configuration was held constant across all subsequent dataset split experiments. We use \texttt{dtype=torch.float32} throughout to ensure numerical precision and gradient stability.

\section{Diagnostic Experiments and Results}
To determine whether the model's performance stems from genuine spatial understanding rather than superficial dataset biases, we divide our findings into three analytical phases. Multiple dataset splitting strategies were evaluated as diagnostic probes. A comprehensive summary of all experimental configurations and their respective outcomes is provided in Table \ref{tab:all_experiments_summary}. Experiment 1 serves as our baseline, where the model was trained on a randomly chosen subset (approximately 78\%) of the original benchmark, achieving a perfect 100.00\% accuracy. While this confirms that the 1B parameter scale possesses the raw capacity for the task, the subsequent experiments systematically ablate this baseline to reveal the syntactic overfitting driving this performance.

\subsection{Phase 1: Capacity vs. Shortcut Learning}
The initial phase investigates whether the model learns true spatial representations or relies on structural pattern-matching heuristics.

In Experiment 2, the model was trained exclusively on T1 and evaluated on the remaining unseen templates (T2 through T6). While it achieved near-perfect accuracy on T2, T3, and T4, it achieved 0\% accuracy on T5 and T6. T1 establishes a spatial rule for an ``inner'' object (a lake), whereas T5 and T6 introduce a structural inversion (an ``outer'' object, the sea). The model blindly applied the T1 heuristic to T5/T6, outputting the exact opposite of the ground truth. This demonstrates severe ``shortcut learning'': while capable of extrapolating multi-step logic from single-step data, the model lacks zero-shot spatial awareness when identical grammatical structures describe inverted spatial configurations.

\subsection{Phase 2: Lexical Ambiguity and Compositional Generalization}
To test how the model resolves conflicting spatial rules, Experiment 3 trained the network on T1, T3, and T5, evaluating it on the remainder. Notably, T1 and T5 utilize the entangled keyword ``shore'' with opposing spatial rules. The model achieved a stratified 61.01\% accuracy. While it maintained robust performance on T2 (94.06\%) and moderate performance on T6 (75.00\%), it scored only 13.96\% on T4. In Experiment 2, the model solved T4 accidentally via a naive inversion shortcut; however, introducing conflicting inner/outer rules in Experiment 3 forced the model to abandon this universal inversion heuristic, causing its zero-shot extrapolation to the unseen T4 syntax to collapse.

Experiment 5 further corroborated this fragility by training the model on the 7/8ths of the entire dataset excluding all instances where the target answer was one particular direction out of the eight (e.g. all instances where the answer was north). Evaluated on the withheld 720 questions, the model achieved 93.89\% accuracy, with all 44 failures occurring exclusively within T3. Because T3 contains the unambiguous keyword ``side,'' the model relied on a shallow pattern-matching shortcut, unlike T1/T5 which forced deep spatial representations.

However, when unambiguous semantic components are provided, the model exhibits true compositional generalization. In Experiment 4, the model was trained on all templates except T4. Evaluated strictly on the withheld T4 structure, it achieved 100.00\% accuracy. Because Template T3 already established the unambiguous syntactic mapping for ``along the [direction] side of,'' the model successfully matched this learned syntactic pattern in T4 while ignoring the trailing axial clause.

\subsection{Phase 3: Semantic Noise vs. Spatial Operands}
The final phase separates the model's understanding of superficial grammar from its reliance on fundamental spatial operands. 

Experiments 6 and 7 stratified the training data by preserving 80\% of each locomotion type and person form, respectively. Both achieved 100.00\% evaluation accuracy, confirming that variations in semantic ``flavor'' variables do not impact spatial vectors. Similarly, Experiment 8 proved that reducing data volume to 80\% across all cardinal directions preserves 100.00\% accuracy, provided no templates are entirely withheld.

We pushed this abstraction further in Experiments 9 and 10 by entirely excluding specific modes of locomotion (``unicycling'', ``perambulating'') and a specific grammatical subject (``She is'') from training. Evaluated exclusively on these unobserved words, the model achieved flawless 100.00\% accuracy. The SLM successfully ignored the unfamiliar vocabulary, relying purely on recognized syntactic patterns to resolve the spatial query. 

Conversely, Experiment 11 completely excluded specific spatial vectors (``north-east'', ``south-west'') from the training prompts. Accuracy collapsed to 33.33\%. The model scored 0\% on T5/T6 by losing positional attention and blindly echoing the prompt's initial heading, and exactly 50\% on T1-T4 by mistakenly applying operations to the heading vector instead of the target vector. This shows that while an SLM can perfectly generalize over missing semantic variables, it cannot organically deduce unobserved spatial operands; it requires a complete spatial vocabulary to prevent regression into shallow positional heuristics.

\begin{table*}[tb]
    \centering
    \begin{tabular}{cllc}
        \toprule
        Exp. & Training Condition & Primary Variable Tested & Eval Acc. \\
        \midrule
        1 & Uniform Baseline ($\sim$78\% of all templates) & Model Parameter Capacity & \textbf{100.00\%} \\
        2 & Trained strictly on T1 & Shortcut Learning & 58.73\% \\
        3 & Mixed conflict (T1, T3, T5) & Lexical Entanglement & 61.01\% \\
        4 & Withheld T4 & Compositional Generalization & \textbf{100.00\%} \\
        5 & Withheld Target Direction & Fragile Heuristics & 93.89\% \\
        6 & Stratified Locomotion (80\% Split) & Data Volume (Flavor) & \textbf{100.00\%} \\
        7 & Stratified Person Form (80\% Split) & Data Volume (Flavor) & \textbf{100.00\%} \\
        8 & Stratified Cardinal Direction (80\% Split) & Data Volume (Spatial) & \textbf{100.00\%} \\
        9 & Withheld Locomotion (OOD verbs) & Semantic Noise Robustness & \textbf{100.00\%} \\
        10 & Withheld Person Form (OOD subject) & Semantic Noise Robustness & \textbf{100.00\%} \\
        11 & Withheld Directions (Missing Spatial Operands) & Spatial Deduction & 33.33\% \\
        \bottomrule
    \end{tabular}
    \caption{Comprehensive summary of the 11 diagnostic ablation experiments. Results demonstrate the strict boundary between the model's robust capability to bypass out-of-distribution (OOD) semantic noise (Exps 9, 10) and its catastrophic failure when deprived of core spatial operands (Exp 11).}
    \label{tab:all_experiments_summary}
\end{table*}

\section{In-Context Learning vs. Fine-Tuning}

A central question in deploying Small Language Models (SLMs) is whether their spatial reasoning deficits can be bridged purely through prompt engineering, or if gradient-based weight updates are strictly necessary. To answer this, the fine-tuning results from our diagnostic probes were compared against an In-Context Learning (ICL) approach using the base \texttt{gemma-3-1b-it} model with no fine-tuning under identical data constraints.

\begin{figure*}[ht]
    \centering
    \includegraphics[width=\textwidth]{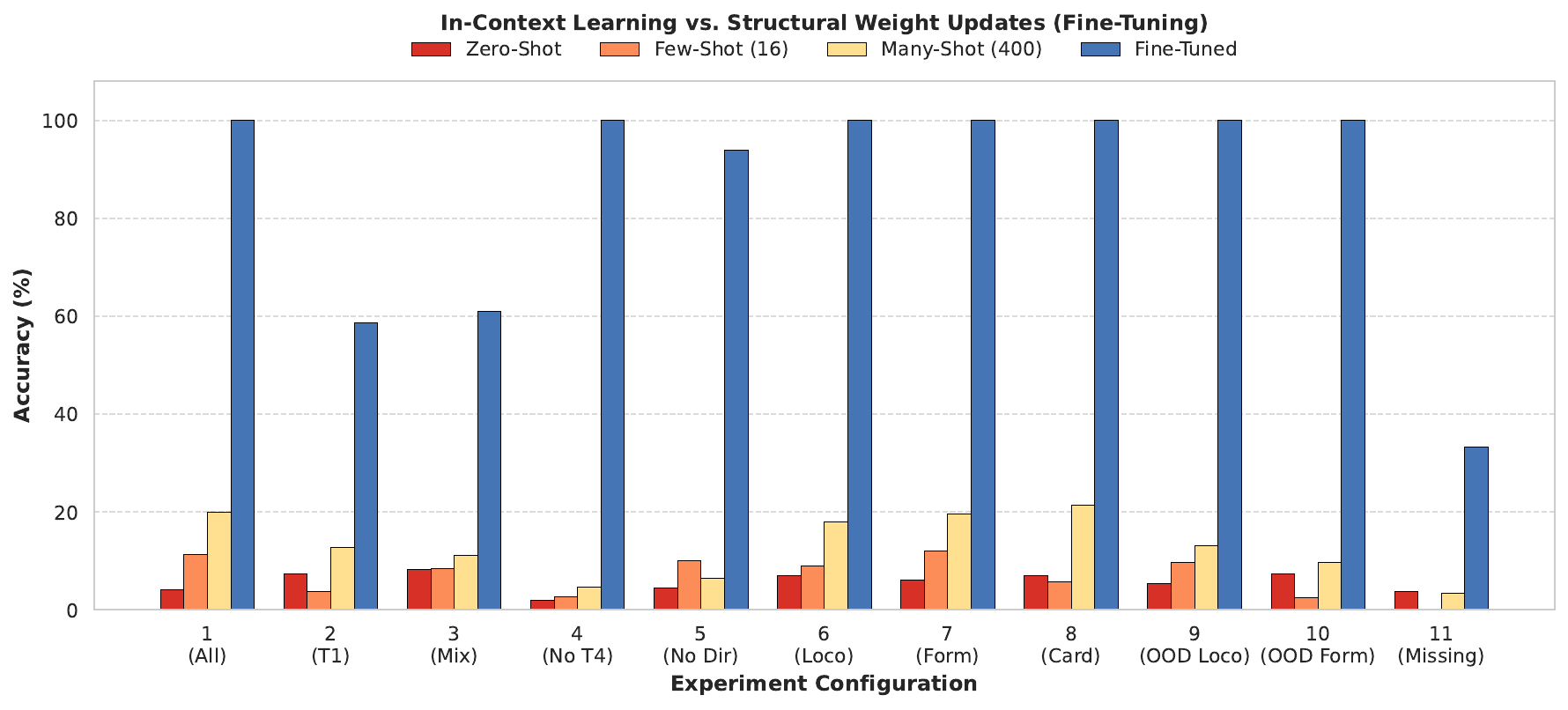}
    \caption{Performance comparison across all 11 diagnostic ablation configurations. In-Context Learning (Zero, Few, and Many-Shot) consistently fails to surpass the $\sim$20\% threshold, while structural fine-tuning deterministically masters the spatial logic in most scenarios.}
    \label{fig:icl_lines}
\end{figure*}

\subsection{Experimental Setup for ICL}
To ensure a strictly controlled comparison, the base model was evaluated against the exact same holdout datasets used for the fine-tuned models. The ICL experiments were conducted across three scaling paradigms:
\begin{itemize}
    \item \textbf{Zero-Shot:} The base model was queried with no prior examples.
    \item \textbf{Few-Shot (16-Shot):} The model was provided a conversational history of 16 random, perfectly balanced examples (exactly 2 examples for each of the 8 target answer directions) extracted directly from the corresponding experiment's training dataset.
    \item \textbf{Many-Shot (400-Shot):} The context window was pushed to its practical limits with 400 balanced examples (50 per target answer direction) from the training set, consuming roughly 14,000 tokens of context.
\end{itemize}

\begin{table*}[tb]
    \centering
    \begin{tabular}{clcccc}
        \toprule
        Exp. & Evaluation Target & Zero-Shot & Few-Shot ICL & Many-Shot ICL & Fine-Tuned \\
        \midrule
        1 & Entire Benchmark & 4.38\% & 11.43\% & 20.00\% & \textbf{100.00\%} \\
        2 & Trained strictly on T1 & 7.35\% & 3.75\% & 12.79\% & 58.73\% \\
        3 & Mixed conflict (T1, T3, T5) & 8.33\% & 8.51\% & 11.08\% & 61.01\% \\
        4 & Withheld T4 & 1.98\% & 2.71\% & 4.69\% & \textbf{100.00\%} \\
        5 & Withheld Target Direction & 4.58\% & 10.14\% & 6.53\% & 93.89\% \\
        6 & Stratified Locomotion & 7.04\% & 8.96\% & 18.09\% & \textbf{100.00\%} \\
        7 & Stratified Person Form & 6.08\% & 11.98\% & 19.62\% & \textbf{100.00\%} \\
        8 & Stratified Cardinal Direction & 6.94\% & 5.73\% & 21.44\% & \textbf{100.00\%} \\
        9 & Withheld Locomotion (OOD) & 5.38\% & 9.81\% & 13.19\% & \textbf{100.00\%} \\
        10 & Withheld Person Form (OOD) & 7.40\% & 2.60\% & 9.79\% & \textbf{100.00\%} \\
        11 & Withheld Directions (Missing Spatial) & 3.82\% & 0.21\% & 3.47\% & 33.33\% \\
        \bottomrule
    \end{tabular}
    \caption{Comprehensive comparison of Base Zero-Shot, Few-Shot In-Context Learning (16 examples), Many-Shot In-Context Learning (400 examples), and Fine-Tuned performance across all 11 diagnostic configurations. All models were evaluated on identical holdout sets.}
    \label{tab:icl_comparison}
\end{table*}

A comprehensive visualization of these diverging paradigms—showing the failure of prompt scaling alongside the deterministic success of fine-tuning—is presented in Figure \ref{fig:icl_lines}.

\subsection{Zero-Shot: The Lexical Echoing Phenomenon}

\begin{figure}[tb]
    \centering
    \includegraphics[width=\columnwidth]{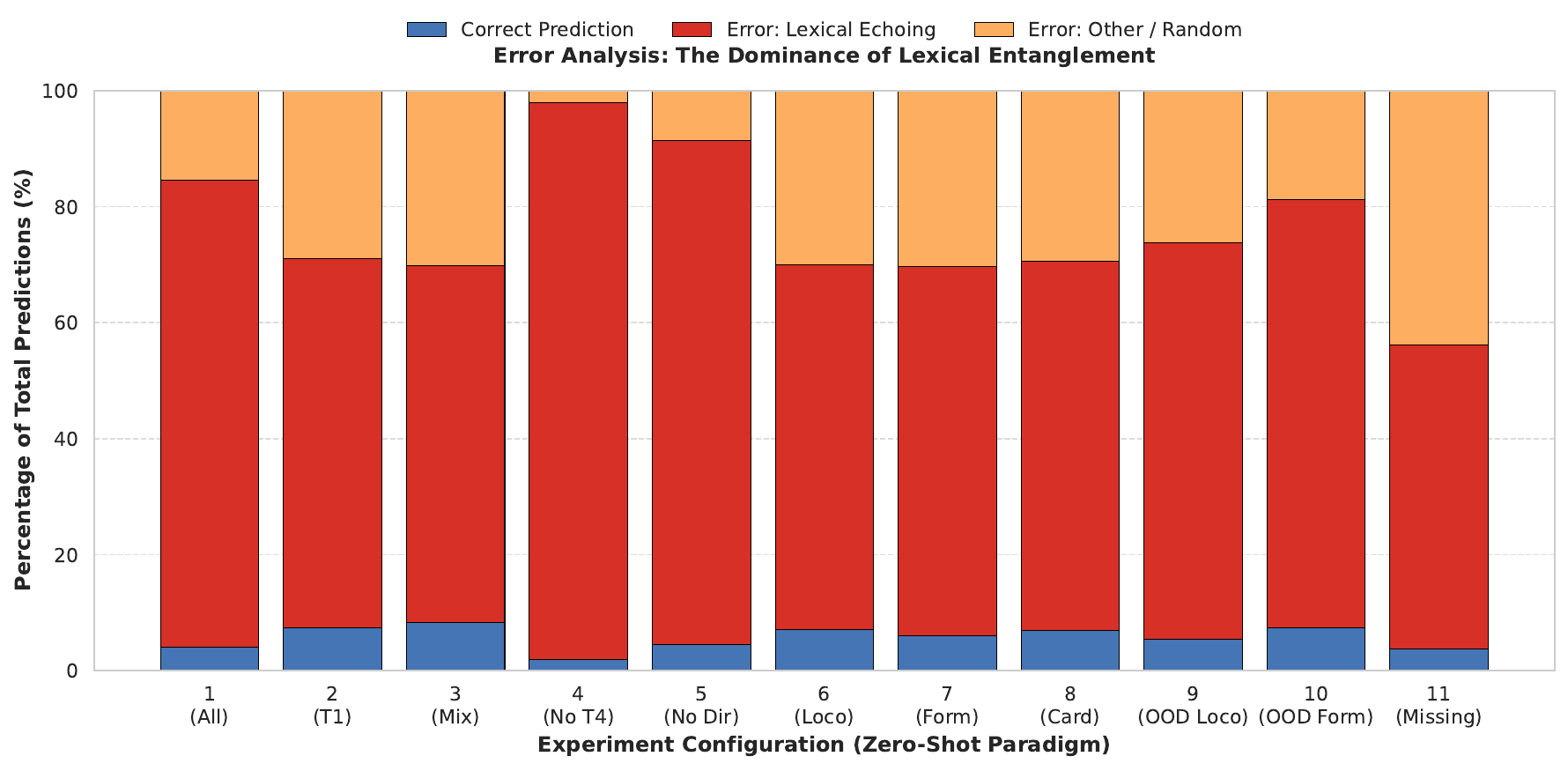}
    \caption{Distribution of predictions in the Zero-Shot paradigm across all 11 experiments. The analysis reveals that the vast majority of failures are not random guesses, but rather instances of \textit{lexical echoing}—where the model shallowly repeats directional tokens found in the prompt rather than computing the spatial vectors.}
    \label{fig:error_analysis}
\end{figure}

When evaluated on the entire, unfiltered benchmark consisting of all 5,760 questions, the base model achieved a global zero-shot accuracy of 4.38\%, which falls significantly below random guessing (12.5\%). As detailed in Table \ref{tab:icl_comparison}, performance on the strictly isolated subsets remained equally dismal, ranging between 1.98\% and 8.33\%.

Qualitative error analysis of the generated outputs revealed that the zero-shot model operates as a shallow linguistic parser rather than a spatial reasoner, a systemic failure mode quantitatively visualized in Figure \ref{fig:error_analysis}. The model suffered from severe \textit{lexical echoing}: 71.62\% of its incorrect predictions simply repeated a cardinal direction explicitly written in the prompt (ranging from 54.44\% in Experiment~11 to 97.87\% in Experiment~4). Furthermore, it demonstrated a strict ``locomotion anchor bias'', overwhelmingly echoing the first direction it encountered in the sentence while ignoring the subsequent structural constraints. This confirms that without intervention, SLMs fall back on spurious linguistic correlations to resolve spatial queries.

\subsection{Few-Shot: Format Learning vs. Negative Transfer}
The transition to a 16-shot ICL paradigm yielded highly bifurcated results. In experiments where accuracy increased (e.g., Experiments 1, 5, 7, and 9), qualitative analysis showed that the prompt successfully taught the model the \textit{format} of the task. The model stopped echoing words verbatim and began adhering to the required lowercase, single-word output constraints.

However, in structurally complex configurations, ICL actively harmed the model's reasoning capabilities, triggering severe \textit{negative transfer}. In Experiment 2, the 16-shot prompt consisted entirely of Template 1 examples. Consequently, the model learned a rigid structural shortcut. When evaluated on Templates 2-6—which invert this spatial logic—the model confidently applied the incorrect shortcut, causing accuracy to drop from 7.35\% (Zero-Shot) to 3.75\% (Few-Shot).

This brittleness was most disastrous in Experiment 11 (Missing Spatial Constraints). The 16-shot prompt trained the model to confidently map complete spatial sentences to specific directions. When evaluated on incomplete spatial constraints, the model's pattern-matching shattered, causing it to hallucinate rigid answers rather than reverting to baseline guessing, resulting in a near-total failure rate of 0.21\%.

\subsection{Many-Shot: The Illusion of Scaling and Mode Collapse}
Expanding the context window to 400 balanced examples generally raised the baseline accuracy, with several experiments plateauing around 20\%. However, an inspection of the raw outputs reveals that this scaling offers merely an illusion of improvement.

Despite being provided a mathematically perfect, uniformly distributed prompt (50 examples per direction), the many-shot model suffered from extreme output mode collapse. Overwhelmed by the massive 14,000-token context, the transformer's attention bottleneck forced the model to anchor onto statistically ``safe'' tokens. For example, in Experiment 2, despite the balanced prompt, the model blindly guessed ``south'' in 3,380 out of 4,800 predictions. In Experiment 3, it defaulted to ``west'' 1,868 times. Scaling up ICL does not impart qualitative spatial reasoning; it merely shifts the model's statistical guessing biases.

\subsection{The Necessity of Weight Updates}
As demonstrated across all 11 configurations, prompt context alone is insufficient to embed robust spatial representations within a 1B-parameter model. ICL relies entirely on shallow, token-level pattern matching across the context window. When faced with lexical entanglement—such as the conflicting spatial relationships governed by the word ``shore'' in Templates 1 and 5—the base model fails to execute spatial deduction.

Conversely, fine-tuning permanently alters the model's parametric memory. The cross-entropy loss penalties incurred during training force the neural network to decouple semantic noise (e.g., verbs, pronouns) from the underlying spatial operands. Ultimately, this comprehensive evaluation proves that while In-Context Learning is a powerful tool for stylistic formatting, true zero-shot compositional generalization in complex spatial and directional reasoning strictly requires the structural weight updates provided by fine-tuning.

\section{The Failure of Generic Spatial Datasets}

Before engineering a highly specialized dataset (the MVD), we investigated whether training the 1B SLM on existing, generic spatial reasoning datasets could organically instill the required spatial logic to solve the \bench{} benchmark. We conducted full-weight fine-tuning (unfreezing all parameters, rather than utilizing low-rank adaptations like LoRA \citep{hu2022lora}) to ensure the model possessed the maximum parametric capacity to construct a generalized spatial cognitive map. 

To evaluate whether generic spatial data can improve upon the base model's 4.38\% zero-shot accuracy, two distinct spatial datasets were utilized:
\begin{enumerate}
    \item \textbf{StepGame:} A spatial reasoning dataset testing multi-hop relations on a 2D plane \citep{li2024stepgame}. We filtered the dataset to include only 1- and 2-hop questions strictly involving cardinal directions.
    \item \textbf{CDC Composition Table:} A matrix explicitly detailing composition of cardinal relations (e.g., if point $x$ is South of point $y$, and $y$ is East of $z$, the model must deduce the spatial relation between $x$ and $z$).
\end{enumerate}

\subsection{The Loss Function and Format Overfitting}
We hypothesized that standard causal language modeling on spatial text leads to format-overfitting rather than genuine logic comprehension. To test this, we applied two distinct loss configurations to the generic datasets: a \textit{Standard Trainer} (which calculates cross-entropy loss over the entire sequence, including the prompt) and a \textit{Custom SFT Trainer} (which applies a -100 mask to the prompt, penalizing the model exclusively for the generated directional output).

The results showed that generic spatial data fails to generalize:
\begin{itemize}
    \item \textbf{StepGame:} Standard training yielded 9.01\% accuracy. Applying SFT prompt-masking improved this slightly to 14.43\%.
    \item \textbf{Composition Table:} Standard training yielded 1.74\% accuracy. Applying SFT prompt-masking yielded 7.22\%.
\end{itemize}

None of these generic datasets came close to solving the spatial task, and the Standard Trainer configurations actively degraded performance, causing the model to generate repetitive conversational text rather than the required single-token directional answers. While the SFT Trainer prevented this format collapse, the spatial logic still failed to transfer to the \bench{} evaluation.

\subsection{Structural Incompatibility}
Qualitative analysis of these failures reveals why generic spatial datasets cannot instill functional navigation logic into SLMs:

\paragraph{The StepGame Disconnect:} StepGame maps abstract points (A to B) on a blank 2D plane. It entirely lacks physical boundaries, topological inversion (interior versus exterior objects), and locomotion prepositions such as ``along,'' which constitute the core mechanical difficulty of the \bench{} benchmark. The semantic noise of StepGame diluted the SLM's attention, failing to teach it the grammatical pattern-matching required for border-based navigation.

\paragraph{The Composition Table Disconnect:} The Composition Table successfully teaches abstract multi-hop spatial reasoning (A $\rightarrow$ B $\rightarrow$ C). However, the \bench{} benchmark never queries multi-point relations; it is strictly a 1-hop cardinal direction mapping problem based on syntactic cues. By training the model on the Composition Table, we inadvertently taught it abstract relational logic. Consequently, when the evaluation script presented the 1-hop syntax of \bench{}, the model failed to find the expected multi-hop variable anchors (points $x$, $y$, $z$) and its logical pathways collapsed entirely, resulting in the 1.74\% failure rate.

These results empirically prove that SLMs do not possess the capacity to absorb generic spatial concepts and dynamically apply them to real-world physical constraints. To successfully inject spatial logic into small models, training must abandon generic datasets and adhere strictly to the structurally analogous syntax of the Minimum Viable Dataset (MVD).

\section{The Minimum Viable Dataset}

The results of the diagnostic ablation study provide insights into the cognitive mechanics of SLMs. While the 1B parameter model possesses the parameter capacity to achieve 100\% accuracy on the dataset, this performance does not stem from true spatial representations. As expected for a text-based model of this scale, its deductions rely entirely on discrete syntactic pattern matching rather than grounded physical logic. 

Consequently, achieving state-of-the-art spatial reasoning on edge devices does not require exhaustive, large-scale datasets with massive semantic diversity. Instead, an SLM will achieve 100\% zero-shot spatial reasoning strictly if its training data satisfies four structural constraints, which we define as the ``Minimum Viable Dataset'' (MVD):

\begin{enumerate}
    \item \textbf{Complete Spatial Vocabulary:} The dataset must contain the exhaustive set of cardinal and intercardinal directions the model will be expected to deduce. Because the model cannot organically infer unobserved spatial directions, omitting specific directions (as seen in Experiment 11) results in the collapse of the model's spatial reasoning pathways and a regression to guessing.
    
    \item \textbf{Intentional Lexical Conflict:} SLMs are highly susceptible to shortcut learning. When trained on simple, non-conflicting rules, the model builds fragile heuristics (e.g., merely outputting the inverse of the prompt's direction). To build deep, robust logical pathways, the dataset must include structural ambiguity---such as the word ``shore'' dictating opposing spatial rules in inner versus outer object configurations---forcing the model to abandon lazy heuristics.
    
    \item \textbf{Atomic Composition:} Every structural rule (e.g., moving along a boundary) and spatial modifier (e.g., turning around) must be observed independently. If the model learns Concept A and Concept B separately, its compositional generalization allows it to successfully combine them to solve entirely novel syntactic structures with perfect accuracy.
    
    \item \textbf{Semantic Minimalism:} The dataset requires almost zero variation in semantic ``flavor'' text. The ablation study demonstrates that a properly fine-tuned model flawlessly identifies and bypasses irrelevant grammatical noise. When presented with completely unobserved verbs (e.g., ``unicycling'') or unseen pronouns, the model effortlessly ignores them to focus purely on the spatial prepositions. 
\end{enumerate}

Ultimately, to embed spatial logic into an SLM, dataset curation should completely abandon semantic variety and focus exclusively on exhaustive spatial completeness and structural conflict. This framework serves as the blueprint for all subsequent experiments and evaluations in this study.

\section{Engineering the Minimum Viable Dataset}

Having established that general spatial datasets fail to transfer to strict cardinal direction reasoning, we constructed a custom training pipeline strictly adhering to the four rules of the Minimum Viable Dataset (MVD) framework. The objective of this phase is not just to achieve 100\% accuracy, but to systematically map the exact syntactic thresholds required for an SLM to generalize spatial rules.

\subsection{Syntactic Template Base and Variable Permutations}
We expanded the core structural mechanics of the original \bench{} templates (specifically focusing on the interior vs. exterior dilemma of T1 and T5) into a robust, generative framework. The base syntactic template was structured as follows:

\begin{quote}
\texttt{[person] [locomotion] [dir1] along the [dir2] [noun] of [subject1]. In which direction is [subject2]?}
\end{quote}

To generate the dataset, we curated 16 distinct subject pairings that replicate the boundary mechanics of a lake or an island (e.g., \textit{border - a blue square - the blue area}, \textit{edge - an oasis - the sand}, \textit{coastline - America - the ocean}). 

Combining these 16 templates with 6 person forms, 10 locomotion verbs, 8 primary directions, and 2 target directions yielded exactly 15,360 questions per dataset split.

\subsection{The Vocabulary and Optimization Axes}
To determine whether the model was learning spatial logic or simply overfitting to vocabulary, every experiment was evaluated across a dual-axis framework:

\paragraph{The Vocabulary Axis:} 
We generated two distinct versions of the training data:
\begin{itemize}
    \item \textbf{In-Distribution (ID) Vocabulary (Overfitting Check):} Utilized the exact person forms (\textit{``You are''}) and locomotion modes (\textit{``walking'', ``driving''}) found in the \bench{} benchmark.
    \item \textbf{Out-of-Distribution (OOD) Vocabulary (Generalization Check):} Utilized completely novel, out-of-distribution vocabulary (e.g., Person forms like \textit{``Linda is'', ``The rover is''}; Locomotion verbs like \textit{``sauntering'', ``traversing''}). This tests the model's true ability to map syntactic relationships regardless of semantic dressing.
\end{itemize}

\paragraph{The Optimization Axis:}
We evaluated two training regimens to isolate how the model processes the question prompt:
\begin{itemize}
    \item \textbf{Standard Trainer:} Calculates full-sequence Cross-Entropy Loss over the entire input sequence, including the prompt.
    \item \textbf{Masked SFT Trainer:} Applies a mathematical mask (\texttt{labels = -100}) to the conversational prompt, penalizing the model exclusively for generating an incorrect directional output.
\end{itemize}

\subsection{The Prompt Consistency Experiments}

\begin{figure}[tb]
    \centering
    \includegraphics[width=\columnwidth]{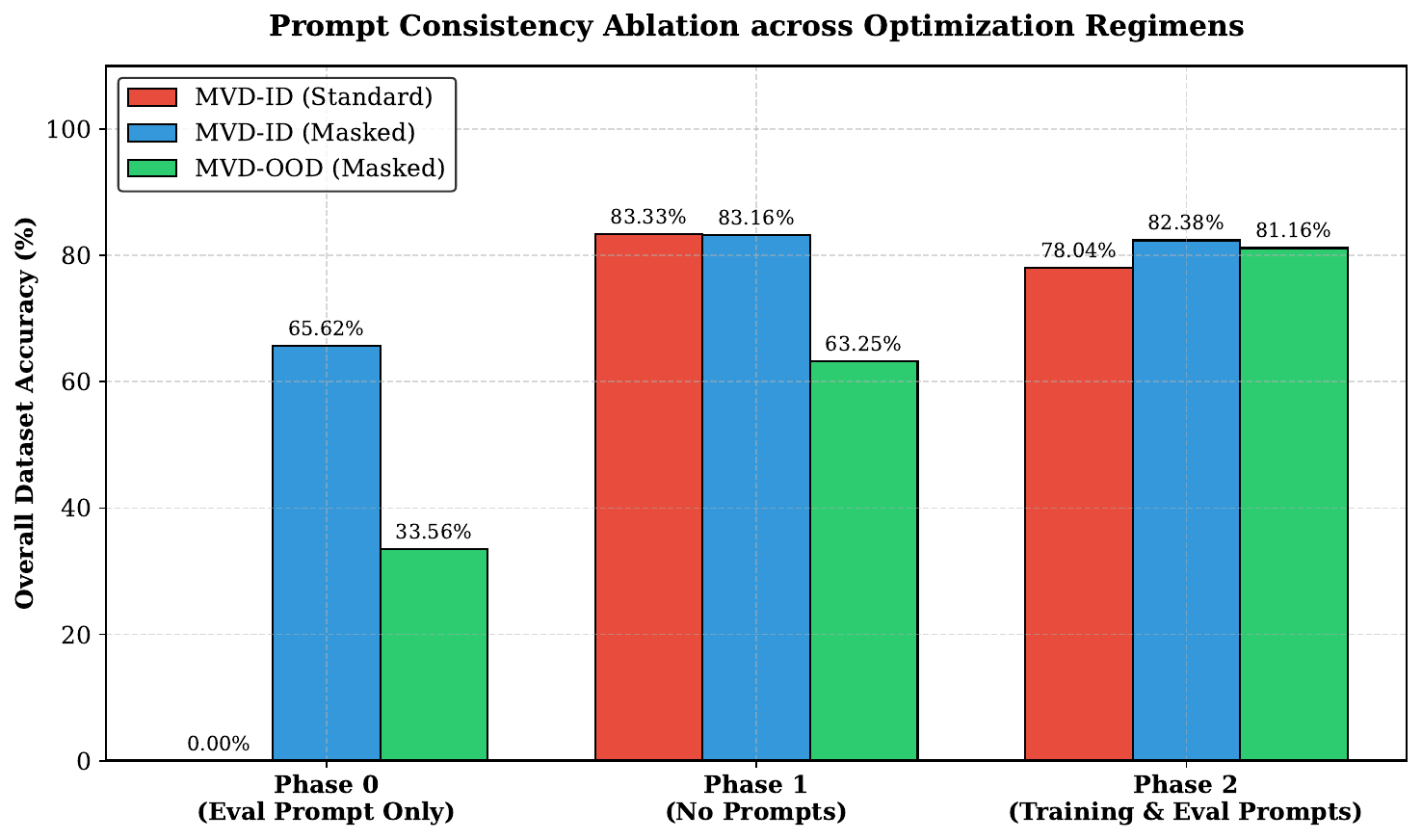}
    \caption{Prompt consistency ablation across optimization regimens. The Masked SFT trainer strictly requires symmetrical prompts across training and evaluation (Phase 2) to successfully generalize out-of-distribution vocabulary (\textbf{MVD-OOD}).}
    \label{fig:prompt_consistency}
\end{figure}

Before addressing complex syntactic structures, we conducted a three-phase experiment using the T1/T5 base structures to determine how prompt consistency affects SLM deduction. Because the base architecture (\texttt{gemma-3-1b-it}) is an instruction-tuned model, stripping away its expected conversational formatting during training introduces a severe distribution shift, which can actively disrupt its reasoning pathways. To map this dependency, we evaluated the following phases, as visualized in Figure \ref{fig:prompt_consistency}:

\begin{itemize}
    \item \textbf{Phase 0 (No Training Prompt, Prompt in Evaluation):} The model was trained purely on the spatial text without the conversational wrapper. When evaluated using the conversational prompt, the \textbf{MVD-ID (Standard)} model failed catastrophically (0.00\%), while the \textbf{MVD-OOD (Masked)} model achieved only 33.56\%.
    \item \textbf{Phase 1 (No Training Prompt, No Evaluation Prompt):} Removing the conversational wrapper from the evaluation script drastically improved results. Both the \textbf{MVD-ID (Masked)} and \textbf{MVD-ID (Standard)} models achieved 100\% on templates T1 and T5, while the \textbf{MVD-OOD (Masked)} achieved 63.25\% overall accuracy.
    \item \textbf{Phase 2 (Prompt in Training, Prompt in Evaluation):} Injecting the conversational wrapper into both the training and evaluation sets yielded highly robust generalization. The \textbf{MVD-OOD (Masked, Prompted)} model successfully mapped the out-of-distribution vocabulary to achieve 100\% on T1 and T5, yielding an 81.16\% global accuracy across the benchmark.
\end{itemize}

These procedural diagnostics yielded three rules for all subsequent experiments: (1) Conversational prompts must be strictly symmetrical across training and evaluation, (2) the Masked SFT optimization is required to prevent the model from wasting parametric capacity on prompt reconstruction, and (3) we must consistently evaluate the OOD vocabulary variants to guarantee we are mapping generalized syntax rather than memorizing localized vocabulary.

\section{Iterative Syntactic Fixes and Edge Cases}

With the foundational pipeline established, the next phase addressed specific structural complexities within the \bench{} templates. Building on the optimization rules established in Phase 2, all subsequent models exclusively utilized out-of-distribution (OOD) vocabulary and Masked SFT optimization to guarantee that the network was learning spatial relations rather than memorizing vocabulary. The goal was to force the SLM to distinguish between spatially relevant prepositions and irrelevant lexical distractors.

\subsection{Resolving Distractor Phrases (Templates T2, T6, and T4)}
Templates T2 and T6 introduce a trailing distractor phrase: \textit{``...and then turn around to head back in the direction you came from.''} Geometrically, this alters the agent's facing orientation but does not change the spatial vector between the agent's location and the target object. The correct answers for T2 and T6 remain identical to T1 and T5, respectively.

To prevent the model from memorizing this exact string, we avoided direct phrase matching. Instead, we injected semantic paraphrases (e.g., \textit{``makes a U-turn,''} \textit{``pivots 180 degrees,''} \textit{``reverses direction''}) into the training data. The resulting \textbf{MVD-OOD (+T2/T6)} model successfully learned to map these variations, achieving 100\% accuracy on templates T1, T2, T4, T5, and T6.

This semantic generalization organically solved Template T4. T4 contains its own axial distractor: \textit{``which runs [dir1] to [opposite\_dir1]''}. Without explicitly training the model on T4 data, the network leveraged its prior training on T2/T6 distractors to safely ignore the T4 trailing clause, achieving 100\% accuracy on T4 organically.

\subsection{The Syntactic Wall (Template T3)}
Template T3 presented the most severe structural challenge, scoring 0.00\% during the \textbf{MVD-OOD (+T2/T6)} evaluation. T3 introduces highly complex internal syntax, injecting path modifiers (\textit{``middle of''}) and explicit locators (\textit{``located in the centre of''}) into the base structure.

To create variations that prevented overfitting, we generated 5 paraphrases for each of the three internal clauses (e.g., \textit{``midpoint of,''} \textit{``gazebo,''} \textit{``situated right at the center of''}). However, combining these variations across the baseline parameters yielded a combinatorial explosion: $5^3 \times 92,160 = 11,520,000$ potential questions. Training on this volume would require an estimated 208 days on our hardware constraint.

To bypass this syntactic wall, we utilized random sampling to select exactly one path modifier, one landmark, and one locator per constraint, generating a distilled dataset of 107,520 examples (requiring approximately 47 hours of training). The resulting model, \textbf{MVD-OOD (+T3)}, achieved a near-perfect overall accuracy of 99.91\%, successfully mapping the complex syntax.

\subsection{The ``Unicycling'' Anomaly}

While the T3 semantic patch brought the overall accuracy to 99.91\%, the model experienced exactly 5 failures out of the 5,760 evaluation questions. These failures exclusively occurred within Template T4, which had previously scored 100\%.

An analysis of the specific failures (IDs 3430, 3590, 3670, 3750, and 3830) revealed an insight into SLM cognitive mechanics. All five prompts followed an identical structure:

\begin{quote}
\textit{[Pronoun] [unicycling] south-east along the south-west side of a road which runs south-east to north-west.}
\end{quote}

In every instance, the ground truth was \textit{north-east}, but the model incorrectly output \textit{north-west}. The model failed \textit{exclusively} when the rare, out-of-distribution token ``unicycling'' was combined with a heavily hyphenated direction (south-east/south-west) and the unseen T4 trailing distractor. 

Teaching the model the highly complex T3 syntax stretched its attention heads. When confronted with unfamiliar vocabulary (unicycling) combined with a distractor clause it had never formally trained on, its pattern-matching failed. This serves as empirical proof of our core thesis: SLMs do not map geographic space. Instead, they rely on fragile surface pattern-matching over familiar sentence templates and vocabulary. When that familiar template is disrupted by novel semantic tokens and unseen distractor clauses, their spatial reasoning collapses.

\subsection{The Ultimate Dataset (MVD)}

\begin{figure*}[ht]
    \centering
    \includegraphics[width=\textwidth]{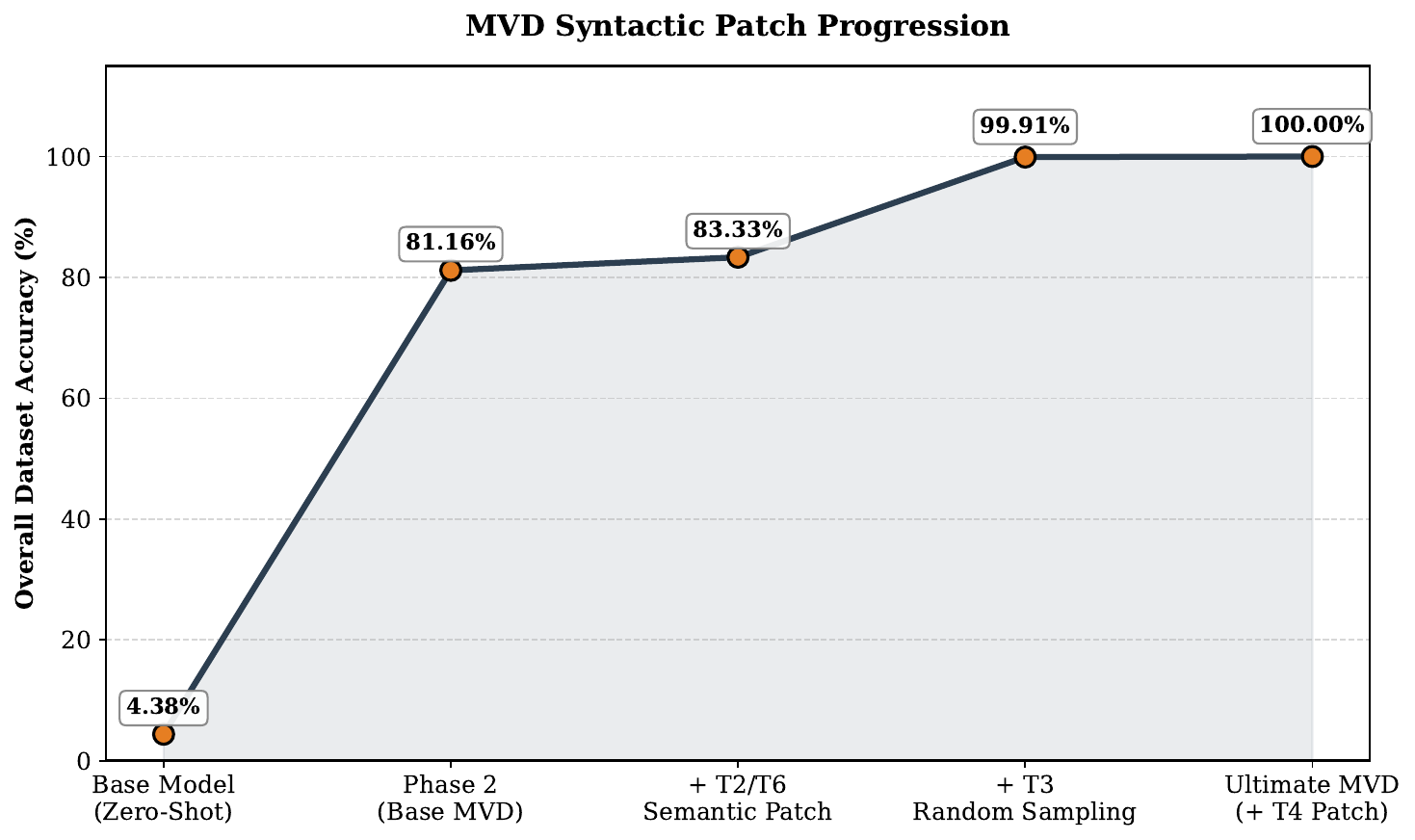}
    \caption{The MVD Syntactic Patch Progression. The chart illustrates the systematic reverse-engineering of the SLM's spatial logic, beginning from a true zero-shot baseline of 4.38\% and culminating in a flawless 100.00\% zero-shot accuracy through targeted semantic paraphrasing.}
    \label{fig:mvd_progression}
\end{figure*}

To formally map the final remaining syntax, we integrated T4 distractor paraphrases (e.g., \textit{``stretches from [dir1] to [dir2],''} \textit{``oriented along a [dir1]-[dir2] axis''}) into the training data. A comprehensive set of generated examples from the Minimum Viable Dataset (MVD), spanning all six template categories, is provided in Appendix \ref{app:mvd_examples}. 

As visualized in the progression chart in Figure \ref{fig:mvd_progression}, by formally mapping every distinct syntactic variation through targeted semantic paraphrasing, the final \textbf{MVD-OOD (Ultimate)} model achieved a deterministic \textbf{100.00\% accuracy} across all 5,760 questions and all 6 templates. 

Importantly, this fine-tuning regime induced strictly positive transfer. Compared to the base model's dismal 4.38\% accuracy, no template scored worse after structural fine-tuning. By systematically reverse-engineering the model's failures, we demonstrate that flawless spatial logic can be embedded into edge-deployable SLMs purely through rigorous syntactic curation.

\section{Architecture Agnosticism: Generalizing the MVD}
\label{sec:agnosticism}

To definitively prove that the success of the Minimum Viable Dataset (MVD) framework is not an artifact of the \texttt{gemma-3-1b-it} architecture, we extended our evaluation to three additional SLMs across different model families: Meta's Llama 3.2 (1B) \citep{grattafiori2024llama}, and Alibaba's Qwen 3.5 (0.8B and 2B) \citep{qwen35}.

As detailed in Table \ref{tab:extra_models}, all three models exhibited the same baseline spatial reasoning deficit, with zero-shot accuracies hovering between 13.00\% and 21.88\%. However, after undergoing the exact same MVD structural fine-tuning regimen, all three architectures achieved a flawless 100.00\% accuracy across all six syntactic templates.

\begin{table}[ht]
    \centering
    \resizebox{\columnwidth}{!}{%
    \begin{tabular}{lccc}
        \toprule
        Model & Params & Base Zero-Shot & MVD Fine-Tuned \\
        \midrule
        Qwen 3.5 & 2B & 17.92\% & \textbf{100.00\%} \\
        Meta Llama 3.2 & 1B & 21.88\% & \textbf{100.00\%} \\
        Qwen 3.5 & 0.8B & 13.00\% & \textbf{100.00\%} \\
        \bottomrule
    \end{tabular}%
    }
    \caption{Validation of architecture agnosticism. Across diverse model families and parameter scales, the MVD framework deterministically elevates near-random baseline spatial performance to perfect accuracy.}
    \label{tab:extra_models}
\end{table}

This deterministic convergence to 100\% across multiple model families confirms that the syntactic constraints defined by the MVD are more broadly applicable across distinct SLM architectures. It demonstrates that the acquisition of spatial logic is fundamentally a matter of training data structure rather than an architectural idiosyncrasy of a specific model family.

\section{Discussion: Reasoning vs. Pattern Matching}

Our empirical findings---specifically the catastrophic failure of generic spatial datasets, the necessity of rigid prompt consistency, and the vulnerability to out-of-distribution semantic noise (the ``Unicycling'' anomaly)---strongly suggest that Small Language Models do not possess an intrinsic cognitive map of physical space---an open question even for LLMs, whose stronger performance may still stem from greater syntactic variety and general reasoning rather than an explicit spatial model. Instead, the SLMs' performance is driven entirely by sophisticated, localized pattern matching. 

As highlighted by \cite{du2022shortcut}, language models frequently engage in \textit{shortcut learning}, exploiting dataset artifacts and superficial syntactic correlations rather than acquiring robust, generalizable understanding. In the context of spatial navigation, the model leverages syntactic boundaries (e.g., the position of prepositions and verbs) as a shortcut proxy for spatial logic. When the syntactic weight becomes unbalanced by novel semantic tokens, the pattern matching shatters, revealing that the model's ``reasoning'' is merely an illusion sustained by a tightly controlled grammatical environment.

While supervised fine-tuning (SFT) on curated syntactic templates is sufficient to solve bounded tasks like the \bench{} benchmark, it is insufficient for instilling generalizable, open-world spatial reasoning across arbitrary tasks (such as multi-hop relational queries). The field must pivot towards more advanced reasoning and alignment frameworks. For example, recent developments in domain-specific SLMs indicate that moving beyond standard fine-tuning to incorporate targeted data generation and reinforcement learning techniques---such as Direct Preference Optimization (DPO)---is strictly necessary to achieve robust domain generalization without shortcut regressions \citep{kumar2025building}. 

Furthermore, achieving zero-shot spatial awareness in edge-deployable autonomous agents will likely require the integration of neurosymbolic architectures or the enforcement of explicit mathematical intermediate steps via Chain-of-Thought (CoT) prompting. These advanced techniques are required to force models to ground their textual parsing in actual spatial representations, rather than allowing them to rely on the fragile illusion of syntactic pattern matching.

\section{Comparison with Large-Scale Language Models}

\begin{figure}[tb]
    \centering
    \includegraphics[width=\columnwidth]{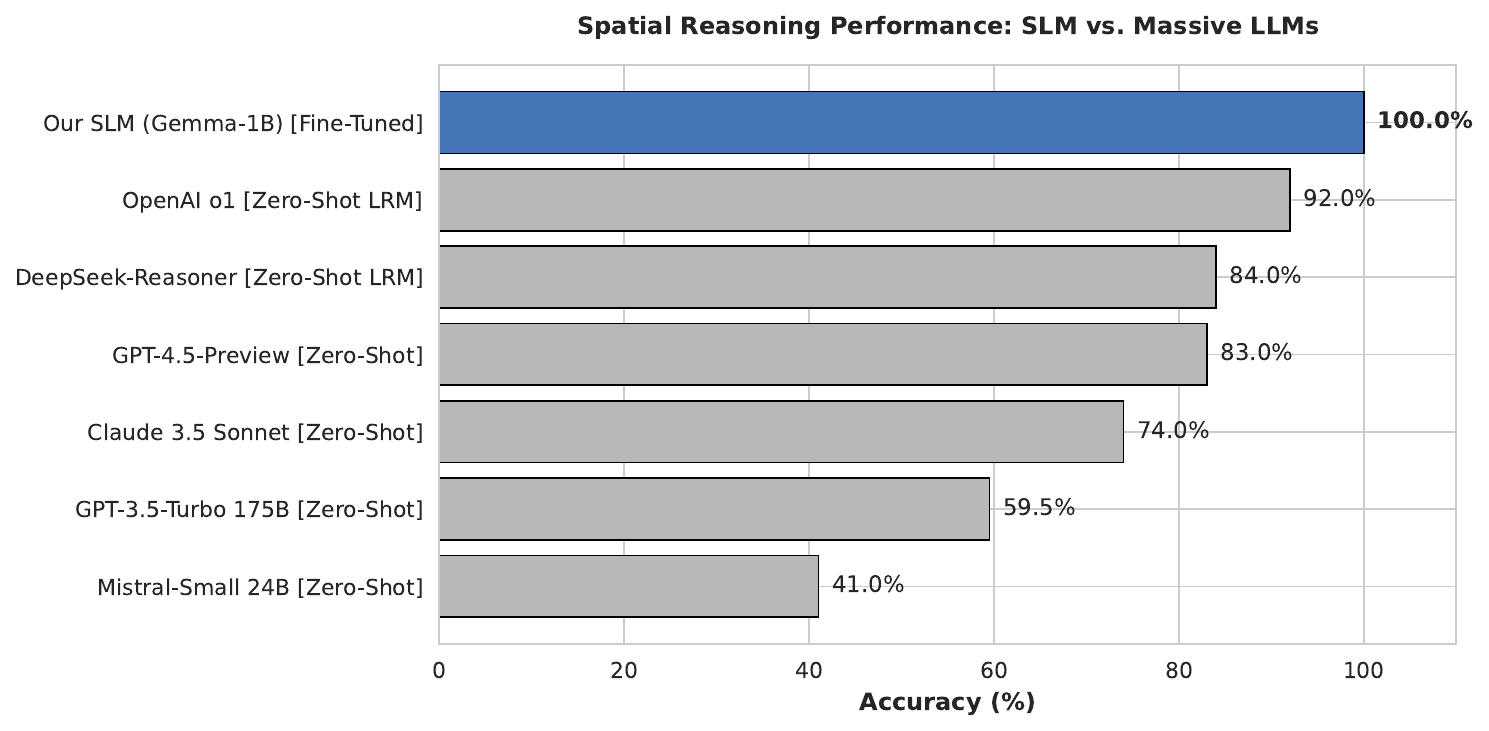}
    \caption{Performance comparison on the \bench{} spatial reasoning benchmark. The fine-tuned 1B model achieves perfect accuracy, outperforming massive proprietary reasoning models. Accuracies for all other models are sourced directly from baseline evaluations by \protect\cite{cohn2024evaluating, cohn2025evaluating}.}
    \label{fig:llm_comparison}
\end{figure}

To fully contextualize the efficacy of this targeted fine-tuning approach, the performance of the base and fine-tuned \texttt{gemma-3-1b-it} models are compared against the zero-shot capabilities of massive Large Language Models and Large Reasoning Models (LRMs) evaluated by \citet{cohn2024evaluating, cohn2025evaluating}. The baseline performances for the massive proprietary models are derived directly from the exhaustive evaluations conducted by \cite{cohn2024evaluating,cohn2025evaluating} on the identical \bench{} large dataset. 

\begin{table*}[tb]
    \centering
    \begin{tabular}{lllc}
        \toprule
        Model & Params & Paradigm & Accuracy \\
        \midrule
        \textbf{Our SLM (Gemma MVD Ultimate)} & \textbf{1B} & \textbf{Fine-Tuned} & \textbf{100.00\%} \\
        \midrule
        OpenAI o1 \citep{cohn2025evaluating} & Massive & Zero-Shot (LRM) & 92.00\% \\
        DeepSeek-Reasoner \citep{cohn2025evaluating} & Massive & Zero-Shot (LRM) & 84.00\% \\
        GPT-4.5-Preview \citep{cohn2025evaluating} & Massive & Zero-Shot & 83.00\% \\
        Claude 3.5 Sonnet \citep{cohn2025evaluating} & Massive & Zero-Shot & 74.00\% \\
        GPT-3.5-Turbo \citep{cohn2024evaluating} & 175B & Zero-Shot & 59.50\% \\
        Mistral-Small \citep{cohn2025evaluating} & 24B & Zero-Shot & 41.00\% \\
        \midrule
        \textbf{Base SLM (Gemma)} & \textbf{1B} & \textbf{Zero-Shot} & \textbf{4.38\%} \\
        \bottomrule
    \end{tabular}
    \caption{Performance comparison on the \bench{} spatial reasoning benchmark. By mastering the underlying syntactic structures of the dataset, the fine-tuned 1B model improves upon its 4.38\% zero-shot baseline to achieve perfect accuracy, outperforming models hundreds of times its size.}
    \label{tab:llm_comparison}
\end{table*}

\subsection{Analysis of the Performance Gap}

As detailed in Table \ref{tab:llm_comparison} and visualized in Figure \ref{fig:llm_comparison}, the pretrained 1B SLM baseline exhibits a severe lack of spatial reasoning capability, scoring just 4.38\%. However, post-MVD fine-tuning, the SLM achieves a deterministic 100.00\% accuracy on out-of-distribution evaluation data, effectively closing the capability gap observed in generalist models.

The highest performing model, OpenAI's \texttt{o1}, relies on massive inference-time compute (chain-of-thought) to achieve a 92.00\% accuracy \citep{cohn2025evaluating}. However, this brute-force approach is computationally expensive and still fails to yield deterministic reliability across all syntactic templates. Cohn and Blackwell observed that LRMs consume significantly more completion tokens when attempting to calculate intercardinal directions, acting as a proxy for the increased cognitive effort required to process multi-axis spatial operands \citep{cohn2025evaluating}.

By contrast, our fine-tuned SLM circumvents this spatial processing bottleneck entirely. Because the SLM's weights have been explicitly updated to map the discrete syntax of the \bench{} benchmark, it does not dynamically ``reason'' through the spatial relations via token generation. It executes the calculation natively as a recognized syntactic pattern. 

\subsection{Parameter Efficiency and Edge Viability}

These results strongly corroborate the hypothesis that task-specific structural optimization fundamentally outperforms scale-based generalization. When comparing the fine-tuned 1B model to the best locally-hostable model from the 2025 baseline (\texttt{Mistral-small-24b}, which achieved only 41.00\% accuracy), the paradigm shift is evident. 

Relying on massive parameter counts to organically deduce spatial logic is highly inefficient and prone to grammatical hallucination. By explicitly isolating the necessary spatial operands and resolving structural conflicts during MVD fine-tuning, an SLM with 24 times fewer parameters can achieve a 59\% absolute increase in accuracy. This establishes a definitive pathway for deploying flawless, domain-specific spatial inference on resource-constrained edge devices without relying on costly, cloud-based LRMs.

Furthermore, as detailed in Appendix~\ref{app:forgetting}, this task-specific fine-tuning preserves the model's core commonsense and scientific reasoning capabilities without catastrophic forgetting.

\section{Conclusion and Future Work}

This work demonstrates that Small Language Models (SLMs) possess the parameter capacity to achieve flawless spatial logic in edge-deployable scenarios, drastically outperforming much larger reasoning models evaluated by \citet{cohn2024evaluating, cohn2025evaluating}. By fine-tuning the 1-billion-parameter Gemma model on the \bench{} benchmark, we achieved a deterministic 100.00\% zero-shot accuracy. This proves that targeted structural optimization supersedes brute-force parameter scaling when navigating physical space. Furthermore, our evaluation confirms that prompt engineering via In-Context Learning (ICL) is insufficient; explicit weight updates are strictly necessary to force the model to resolve lexical entanglement and embed robust spatial heuristics.

However, our ablation studies reveal a fundamental cognitive reality: text-only SLMs act purely as linguistic parsers bounded by their vocabulary. They do not possess a continuous cognitive map of the physical world. While human spatial reasoning relies on mental imagery to visualize actors, trajectories, and boundaries, text-only models rely on the fragile illusion of syntactic pattern matching. Although the Minimum Viable Dataset (MVD) successfully patches these vulnerabilities, the underlying lack of true spatial grounding remains.

Consequently, future work must explore methodologies that bridge this cognitive gap. One promising avenue is the integration of neurosymbolic architectures, coupling the SLM's semantic parsing capabilities with deterministic mathematical engines to ground textual vectors in actual geometric coordinate spaces. Furthermore, overcoming the limitations of standard Supervised Fine-Tuning (SFT) will require advanced alignment techniques. Utilizing Reinforcement Learning (RL), Direct Preference Optimization (DPO) \citep{kumar2025building}, and enforcing explicit mathematical intermediate steps via Chain-of-Thought (CoT) prompting could force models to optimize for robust logical pathways rather than superficial dataset artifacts. 

Alternatively, extending this research to diagrammatic reasoning via Vision-Language Models (VLMs) trained on procedurally generated visual spatial data could instill continuous spatial cognition and organically bypass the need for rigid syntactic curation. Ultimately, embedding true spatial reasoning---whether through neurosymbolic grounding, advanced preference alignment, or multimodal visualization---is a strictly necessary progression toward robust embodied artificial intelligence, where autonomous agents must seamlessly bridge high-level reasoning with physical control systems in the real world.

% =========================================================================================

\appendix

\section{Evaluating Catastrophic Forgetting and General Intelligence Retention}
\label{app:forgetting}

A primary concern when full-parameter fine-tuning a Small Language Model (SLM) on narrow, highly structured spatial distributions is \textit{catastrophic forgetting}. Because a 1-billion-parameter architecture (\texttt{gemma-3-1b-it}) possesses a finite parametric capacity, forcing the network to aggressively learn rigid directional vector mappings over multiple epochs risks overwriting pre-trained weights, thereby degrading general world knowledge and reasoning capabilities.

To rigorously evaluate whether our spatial fine-tuning regime compromised foundational intelligence, we benchmarked the pretrained base model, all 11 diagnostic ablation models (1--11), and a model fine-tuned on the cardinal direction calculus (CDC) composition table \citep{gardelakos2025can} across four independent, out-of-distribution (OOD) benchmarks:
\begin{itemize}
    \item \textbf{PIQA (Physical Interaction Question Answering)} \citep{bisk2020piqa}\textbf{:} Measures physical commonsense reasoning.
    \item \textbf{ARC-Easy \& ARC-Challenge (AI2 Reasoning Challenge)} \citep{clark2018think}\textbf{:} Evaluates grade-school level scientific reasoning across standard and complex questions.
    \item \textbf{StepGame} \citep{shi2022stepgame}\textbf{:} Evaluates multi-hop qualitative spatial relational reasoning.
\end{itemize}

All OOD evaluations were conducted deterministically using the EleutherAI Language Model Evaluation Harness (\texttt{lm-eval}) \citep{gao2023evalharness}. For multiple-choice datasets (PIQA, ARC, StepGame), the harness bypasses probabilistic sampling and computes exact target log-likelihoods, guaranteeing 100\% mathematical reproducibility.

\begin{figure*}[ht]
    \centering
    \includegraphics[width=\textwidth]{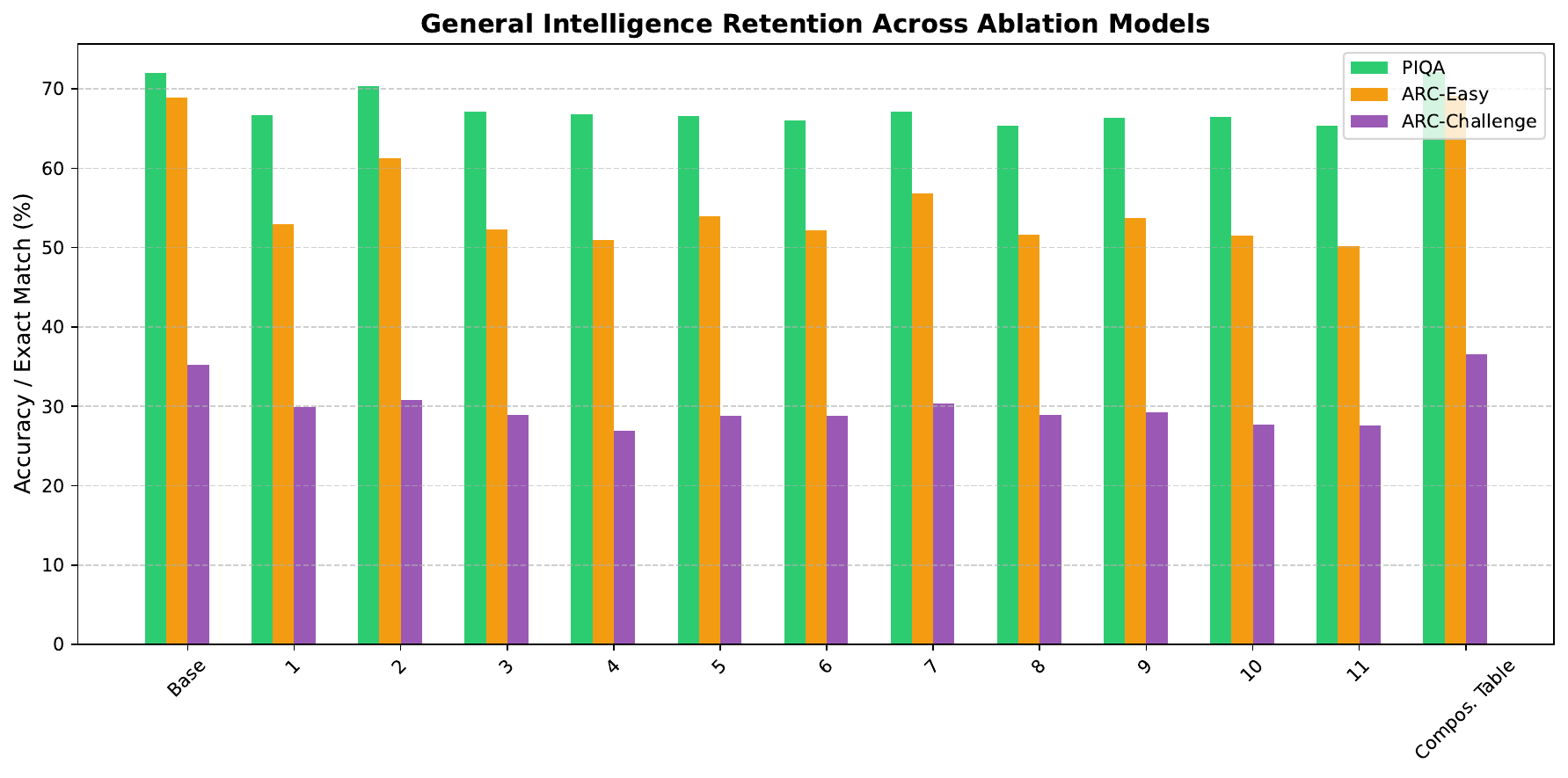}
    \caption{Evaluation of general intelligence retention across the Base model, the 11 diagnostic ablation models (1--11), and the Composition Table model across PIQA, ARC-Easy, and ARC-Challenge.}
    \label{fig:benchmark_retention}
\end{figure*}

\subsection{Commonsense and Scientific Reasoning Retention}

The base \texttt{gemma-3-1b-it} model established zero-shot baselines of 71.98\% on PIQA, 68.90\% on ARC-Easy, and 35.24\% on ARC-Challenge. 

As visually detailed in Figure \ref{fig:benchmark_retention}, fine-tuning on the \bench{} spatial sub-distributions induced only minor, bounded performance shifts across commonsense and scientific benchmarks rather than catastrophic collapse:
\begin{itemize}
    \item \textbf{PIQA Retention:} Across all 11 diagnostic variants, PIQA accuracy remained tightly bounded between 65.34\% (Model 11) and 70.29\% (Model 2), representing a minor degradation of only 1.7\% to 6.6\% from the base model.
    \item \textbf{ARC Resilience:} On ARC-Challenge, performance showed a modest decrease from the baseline, hovering between 26.96\% (Model 4) and 30.80\% (Model 2). ARC-Easy experienced a moderate shift from 68.90\% down to a range of 50.13\%--61.24\%.
\end{itemize}

This minor degradation reflects the standard parametric trade-off of full-parameter fine-tuning on narrow tasks: allocating weights to master localized spatial vector syntax slightly dilutes general physical commonsense. However, because performance remains far above random-chance thresholds, we conclude that the core logical substrate remains intact.

\subsection{Multi-Hop Spatial Generalization: \bench{} vs. Composition Table on StepGame}

To evaluate whether localized fine-tuning instills a generalized spatial cognitive map, all models were evaluated on the StepGame benchmark, which requires multi-hop spatial deduction across complex agent narratives.

\begin{figure}[ht]
    \centering
    \includegraphics[width=\columnwidth]{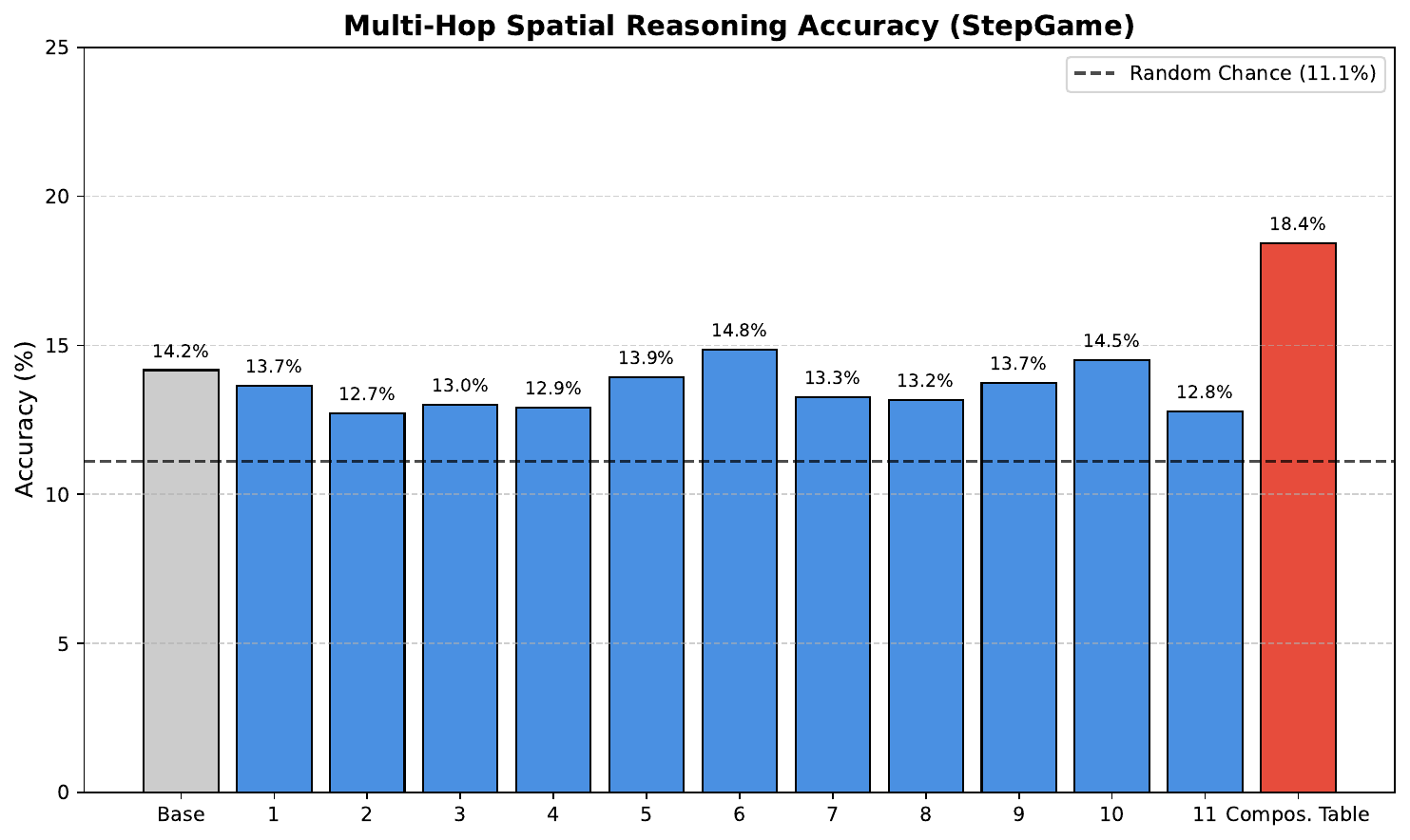}
    \caption{Multi-hop spatial reasoning performance on StepGame across all models. Models fine-tuned exclusively on localized single-hop \bench{} templates (1--11) perform identically to the baseline ($\sim$12.7\%--14.8\%), whereas exposure to the combinatorial Composition Table yields a statistically significant 30\% relative improvement (18.4\%).}
    \label{fig:stepgame_results}
\end{figure}

As shown in Figure \ref{fig:stepgame_results}, models fine-tuned exclusively on localized, single-hop \bench{} templates (Models 1--11) achieved StepGame accuracies ranging between 12.73\% and 14.85\%, scoring statistically identically to the pre-trained Base model (14.17\%) and barely exceeding random chance (11.1\%).

However, training the model on the exhaustive, multi-point relational matrix of the Composition Table dataset successfully induced multi-hop spatial generalization, increasing StepGame accuracy to \textbf{18.43\%}---an absolute gain of 4.26\% (30.1\% relative improvement) over the baseline. 

This contrast highlights the strict limits of the model's pattern-matching behavior. Training on localized, single-step templates (\bench{}) teaches the model to resolve specific natural language prepositions, but provides no structural template for multi-hop queries. Conversely, exposure to the Composition Table trains the model to recognize multi-point chained syntax, yielding a modest structural overlap with StepGame. However, in both cases, the model is not constructing true spatial representations; its deductions remain strictly confined to the specific syntactic structures observed during fine-tuning.

\subsection{The Edge-Deployment Trade-Off}

Ultimately, these evaluations prove that achieving spatial reasoning in an SLM does not require sacrificing the model's core intelligence, provided the formatting constraints are managed. However, even the minor degradations observed in general scientific knowledge (ARC-Easy) are entirely acceptable within the context of our primary motivation: deploying domain-specific agents on resource-constrained edge hardware. 

When engineering autonomous robotics or local navigation agents, the operational priority is flawless, deterministic execution of localized tasks, not generalized encyclopedic knowledge. By proving that 100\% spatial accuracy can be achieved with only a marginal trade-off in broad commonsense, this methodology validates the creation of highly specialized, edge-deployable spatial experts that do not suffer from true catastrophic cognitive collapse.

\section{Minimum Viable Dataset (MVD) Examples}
\label{app:mvd_examples}

This appendix provides representative examples of the questions used to construct the Minimum Viable Dataset (MVD), categorized by their underlying syntactic templates (T1 through T6).

\subsection*{Template T1: Standard Interior Boundary}
Tests basic internal boundaries with no distractors.
\begin{itemize}
    \item \textit{Example 1:} Linda is promenading east along the north border of a blue square. In which direction is the blue square?
    \\ \textbf{Answer:} South
    \item \textit{Example 2:} Jeff is sauntering south along the east border of San Marino. In which direction is San Marino?
    \\ \textbf{Answer:} West
\end{itemize}

\subsection*{Template T2: Interior Boundary with Orientation Distractor}
Adds an action where the subject turns around, testing if the model ignores the change in facing direction.
\begin{itemize}
    \item \textit{Example 1:} Linda is promenading east along the north edge of a ship and turns to face the opposite way, in which direction is the ship?
    \\ \textbf{Answer:} South
    \item \textit{Example 2:} The robot is traversing west along the south coastline of America and then makes a U-turn, in which direction is America?
    \\ \textbf{Answer:} North
\end{itemize}

\subsection*{Template T3: Interior Boundary with Structural Injection}
Adds extra descriptions about midpoints and landmarks to test if the model gets confused by irrelevant details.
\begin{itemize}
    \item \textit{Example 1:} Linda is bounding east along the center point of the north border of a blue square; in which direction is the playground, positioned in the exact middle of the blue square?
    \\ \textbf{Answer:} South
    \item \textit{Example 2:} Jeff is promenading north along the middle portion of the east edge of an iceberg; in which direction is the sculpture, erected dead-center within an iceberg?
    \\ \textbf{Answer:} West
\end{itemize}

\subsection*{Template T4: Interior Boundary with Axial Distractor}
Adds a description of the path's axis to see if the model gets tricked by the extra directional words.
\begin{itemize}
    \item \textit{Example 1:} Linda is gliding east along the north side of a path which trends from east to west. In which direction is the path?
    \\ \textbf{Answer:} South
    \item \textit{Example 2:} The rover is marching south-west along the south-east side of a highway which stretches from south-west to north-east. In which direction is the highway?
    \\ \textbf{Answer:} North-West
\end{itemize}

\subsection*{Template T5: Standard Exterior Boundary}
Tests external boundaries, asking for the direction of the surrounding environment instead of the internal object.
\begin{itemize}
    \item \textit{Example 1:} The dog is crawling south along the east border of San Marino. In which direction is Italy?
    \\ \textbf{Answer:} East
    \item \textit{Example 2:} Jeff is skipping east along the north edge of a ship. In which direction is the ocean?
    \\ \textbf{Answer:} North
\end{itemize}

\subsection*{Template T6: Exterior Boundary with Orientation Distractor}
Combines the external boundary structure with the turning-around distractor from T2.
\begin{itemize}
    \item \textit{Example 1:} The avatar is sprinting south along the east edge of a carpet and turns to face the opposite way, in which direction is the floor?
    \\ \textbf{Answer:} East
    \item \textit{Example 2:} Jeff is bounding east along the north edge of a pool and reverses direction, in which direction is the deck?
    \\ \textbf{Answer:} North
\end{itemize}

\section*{Limitations}
While our results demonstrate that targeted syntactic curation enables 1B-parameter SLMs to achieve deterministic accuracy on the \bench{} benchmark, two primary limitations should be noted:

\begin{itemize}
    \item \textbf{Single-Seed Training Runs:} Due to computational constraints on our single consumer-grade GPU setup, each fine-tuning configuration was executed with a single random seed rather than repeated across multiple runs. Consequently, we do not report standard deviations across training seeds, or even multiple runs on the same seed.
    \item \textbf{Model Scale and Baseline Coverage:} Our fine-tuning experiments focus on compact models between 0.8B and 2B parameters tailored for edge deployment, and our comparisons against large proprietary models rely on the published zero-shot evaluations of \citet{cohn2024evaluating, cohn2025evaluating} rather than re-evaluating more recent frontier releases.
\end{itemize}

\section*{Ethical Statement}
There are no ethical issues.

% References
\bibliographystyle{named}
\bibliography{references} 

@inproceedings{cohn2024evaluating,
  title={Evaluating the Ability of Large Language Models to Reason About Cardinal Directions (Short Paper)},
  author={Cohn, Anthony G and Blackwell, Robert E},
  booktitle={16th International Conference on Spatial Information Theory (COSIT 2024)},
  year={2024},
  organization={Schloss Dagstuhl--Leibniz-Zentrum f{\"u}r Informatik}
}

@article{cohn2025evaluating,
  title={Evaluating the Ability of Large Language Models to Reason about Cardinal Directions, Revisited},
  author={Cohn, Anthony G and Blackwell, Robert E},
  journal={arXiv preprint arXiv:2507.12059},
  year={2025}
}

@misc{belcak2025small,
  title = {Small Language Models are the Future of Agentic AI},
  author = {Belcak, Peter and Heinrich, Greg and Diao, Shizhe and Fu, Yonggan and Dong, Xin and Muralidharan, Saurav and Lin, Yingyan Celine and Molchanov, Pavlo},
  year = {2025},
  eprint = {2506.02153},
  archivePrefix = {arXiv},
  primaryClass = {cs.AI},
  url = {https://arxiv.org/abs/2506.02153},
  doi = {10.48550/arXiv.2506.02153}
}

@article{jhandi2025small,
  title={Small Language Models for Efficient Agentic Tool Calling: Outperforming Large Models with Targeted Fine-tuning},
  author={Jhandi, Polaris and Kazi, Owais and Subramanian, Shreyas and Sendas, Neel},
  journal={arXiv preprint arXiv:2512.15943},
  year={2025}
}

@inproceedings{gardelakos2025can,
  title={Can Large Reasoning Models Reason about Spatial Relations?},
  author={Gardelakos, Eirinaios-Odysseas and Kyriakopoulos, Vasileios and Pantazi, Despina-Athanasia and Kapopoulos, Orestis-Minas and Tsourma, Maria and Koubarakis, Manolis},
  booktitle={Proceedings of the 8th ACM SIGSPATIAL International Workshop on AI for Geographic Knowledge Discovery},
  pages={81--91},
  year={2025}
}

@inproceedings{xie2025evaluating,
  title={Evaluating Intrinsic Geospatial Topological Reasoning in {LLM}s},
  author={Xie, Shaolin and Hsu, Shang-Ling and Zhang, Qihan and Gao, Yiming and Shahabi, Cyrus and Sabek, Ibrahim},
  booktitle={Proceedings of the 1st ACM SIGSPATIAL International Workshop on Generative and Agentic AI for Multi-Modality Space-Time Intelligence},
  pages={43--48},
  year={2025}
}

@article{burnell2023rethink,
  title={Rethink reporting of evaluation results in {AI}},
  author={Burnell, Ryan and Schellaert, Wout and Burden, John and Ullman, Tomer D and Martinez-Plumed, Fernando and Tenenbaum, Joshua B and Rutar, Danaja and Cheke, Lucy G and Sohl-Dickstein, Jascha and Mitchell, Melanie and  Kiela, Douwe  and  Shanahan, Murray and  Voorhees, Ellen M and Cohn, Anthony G. and  Leibo, Joel Z. and   Hernandez-Orallo, Jose},
  journal={Science},
  volume={380},
  number={6641},
  pages={136--138},
  year={2023},
  publisher={American Association for the Advancement of Science}
}

@article{wularge,
  title={Do Large Language Models have Spatial Cognitive Abilities?},
  author={Wu, Ruoling and Guo, Danhuai},
  journal={ACM Transactions on Intelligent Systems and Technology},
  publisher={ACM New York, NY},
year={2025}
}

@inproceedings{yin2025multimodal,
  title={Do multimodal language models really understand direction? {A} benchmark for compass direction reasoning},
  author={Yin, Hang and Lin, Zhifeng and Liu, Xin and Sun, Bin and Li, Kan},
  booktitle={ICASSP 2025-2025 IEEE International Conference on Acoustics, Speech and Signal Processing (ICASSP)},
  pages={1--5},
  year={2025},
  organization={IEEE}
}

@misc{yamada2024evaluating,
      title={Evaluating Spatial Understanding of Large Language Models},
      author={Y Yamada and Y Bao and A K Lampinen and J Kasai and I Yildirim},
      year={2024},
      eprint={2310.14540},
      archivePrefix={arXiv},
      primaryClass={cs.CL}
}

@article{du2022shortcut,
  title={Shortcut Learning of Large Language Models in Natural Language Understanding},
  author={Du, Mengnan and He, Fengxiang and Zou, Na and Tao, Dacheng and Hu, Xia},
  journal={arXiv preprint arXiv:2208.11857},
  year={2022}
}

@inproceedings{kumar2025building,
  title={Building Domain-Specific Small Language Models via Guided Data Generation},
  author={Kumar, Aman and Amin, Ekant Muljibhai and Lee, Xian Yeow and Vidyaratne, Lasitha and Farahat, Ahmed K and Ghosh, Dipanjan D and Koreeda, Yuta and Gupta, Chetan},
  booktitle={Thirty-Eighth Annual Conference on Innovative Applications of Artificial Intelligence (IAAI-26)},
  year={2025}
}

@article{li2024stepgame,
  title={Advancing Spatial Reasoning in Large Language Models: An In-Depth Evaluation and Enhancement Using the {StepGame} Benchmark},
  author={Li, Fangjun and Hogg, David C. and Cohn, Anthony G.},
  journal={arXiv preprint arXiv:2401.03991},
  year={2024}
}

@article{firoozi2023foundation,
  title={Foundation Models in Robotics: Applications, Challenges, and the Future},
  author={Roya Firoozi and Johnathan Tucker and Stephen Tian and Anirudha Majumdar and Jiankai Sun and Weiyu Liu and Yuke Zhu and Shuran Song and Ashish Kapoor and Karol Hausman and Brian Ichter and Danny Driess and Jiajun Wu and Cewu Lu and Mac Schwager},
  journal={arXiv preprint arXiv:2312.07843},
  year={2023}
}

@article{zhou2019edge,
  title={Edge intelligence: Paving the last mile of artificial intelligence with edge computing},
  author={Zhou, Zhi and Chen, Xu and Li, En and Zeng, Liekang and Luo, Ke and Zhang, Junshan},
  journal={Proceedings of the IEEE},
  volume={107},
  number={8},
  pages={1738--1762},
  year={2019},
  publisher={IEEE}
}

@inproceedings{goyal1997,
  author    = {Goyal, Rajal and Egenhofer, Max J.},
  title     = {Similarity of Cardinal Directions},
  booktitle = {The Annual Assembly and the Summer Retreat of University Consortium for Geographic Information Systems Science},
  year      = {1997}
}

@inproceedings{goyal2001,
  author    = {Goyal, Rajal K. and Egenhofer, Max J.},
  title     = {Similarity of Cardinal Directions},
  booktitle = {Advances in Spatial and Temporal Databases (SSTD 2001)},
  year      = {2001},
  address   = {Redondo Beach, CA, USA},
  doi       = {10.1007/3-540-47724-1_3}
}

@article{skiadopoulos2004,
  author    = {Skiadopoulos, Spiros and Koubarakis, Manolis},
  title     = {Composing cardinal direction relations},
  journal   = {Artificial Intelligence},
  year      = {2004},
  doi       = {10.1016/S0004-3702(03)00137-1}
}

@article{gemma2025gemma3,
  author       = {{Gemma Team} and Kamath, Aishwarya and Ferret, Johan and others},
  title        = {Gemma 3 Technical Report},
  journal      = {CoRR},
  volume       = {abs/2503.19786},
  year         = {2025},
  url          = {https://doi.org/10.48550/arXiv.2503.19786},
  doi          = {10.48550/ARXIV.2503.19786},
  eprinttype    = {arXiv},
  eprint       = {2503.19786}
}

@article{openai2023gpt4,
  author       = {{OpenAI}},
  title        = {{GPT-4} Technical Report},
  journal      = {CoRR},
  volume       = {abs/2303.08774},
  year         = {2023},
  eprinttype   = {arXiv},
  eprint       = {2303.08774}
}

@article{touvron2023llama2,
  author       = {Touvron, Hugo and Martin, Louis and Stone, Kevin and others},
  title        = {Llama 2: Open Foundation and Fine-Tuned Chat Models},
  journal      = {CoRR},
  volume       = {abs/2307.09288},
  year         = {2023},
  eprinttype   = {arXiv},
  eprint       = {2307.09288}
}

@article{geminiteam2023gemini,
  author       = {{Gemini Team} and Anil, Rohan and Borgeaud, Sebastian and others},
  title        = {Gemini: A Family of Highly Capable Multimodal Models},
  journal      = {CoRR},
  volume       = {abs/2312.11805},
  year         = {2023},
  eprinttype   = {arXiv},
  eprint       = {2312.11805}
}

@inproceedings{bisk2020piqa,
  author       = {Bisk, Yonatan and Zellers, Rowan and Le Bras, Ronan and Gao, Jianfeng and Choi, Yejin},
  title        = {{PIQA}: Reasoning about Physical Commonsense in Natural Language},
  booktitle    = {Proceedings of the AAAI Conference on Artificial Intelligence},
  volume       = {34},
  pages        = {7432--7439},
  year         = {2020},
  doi          = {10.1609/aaai.v34i05.6239}
}

@article{clark2018think,
  author       = {Clark, Peter and Cowhey, Isaac and Etzioni, Oren and Khot, Tushar and Sabharwal, Ashish and Schoenick, Carissa and Tafjord, Oyvind},
  title        = {Think you have Solved Question Answering? {Try} {ARC}, the {AI2} Reasoning Challenge},
  journal      = {CoRR},
  volume       = {abs/1803.05457},
  year         = {2018},
  eprinttype   = {arXiv},
  eprint       = {1803.05457}
}

@inproceedings{shi2022stepgame,
  author       = {Shi, Zhengxiang and Zhang, Qiang and Lipani, Aldo},
  title        = {{StepGame}: A New Benchmark for Robust Multi-Hop Spatial Reasoning in Texts},
  booktitle    = {Proceedings of the AAAI Conference on Artificial Intelligence},
  volume       = {36},
  pages        = {11321--11329},
  year         = {2022},
  doi          = {10.1609/aaai.v36i10.21383}
}

@misc{gao2023evalharness,
  author       = {Gao, Leo and Tow, Jonathan and Abbasi, Baber and Biderman, Stella and others},
  title        = {A framework for few-shot language model evaluation},
  year         = {2023},
  publisher    = {Zenodo},
  version      = {v0.4.0},
  doi          = {10.5281/zenodo.10256836},
  url          = {https://zenodo.org/records/10256836}
}

@article{grattafiori2024llama,
  author       = {Grattafiori, Aaron and Dubey, Abhimanyu and Jauhri, Abhinav and others},
  title        = {The {Llama} 3 Herd of Models},
  journal      = {CoRR},
  volume       = {abs/2407.21783},
  year         = {2024},
  eprinttype   = {arXiv},
  eprint       = {2407.21783}
}

@misc{qwen35,
  author       = {{Qwen Team}},
  title        = {{Qwen3.5}: Towards Native Multimodal Agents},
  year         = {2026},
  url          = {https://qwen.ai/blog?id=qwen3.5}
}

@inproceedings{hu2022lora,
  author       = {Hu, Edward J. and Shen, Yelong and Wallis, Phillip and Allen-Zhu, Zeyuan and Li, Yuanzhi and Wang, Shean and Wang, Lu and Chen, Weizhu},
  title        = {{LoRA}: Low-Rank Adaptation of Large Language Models},
  booktitle    = {International Conference on Learning Representations (ICLR)},
  year         = {2022},
  url          = {https://openreview.net/forum?id=nZeVKeeFYf9}
}

\end{document}